\documentclass[journal]{IEEEtran}
\usepackage{graphicx}
\usepackage{subcaption}
\usepackage{amsmath,amssymb} % define this before the line numbering.
\usepackage{color}
\usepackage{tabu}
\usepackage{multirow}

\begin{document}

\title{Synthetic data generation for end-to-end thermal infrared tracking}
%\title{Data hallucination for end-to-end thermal infrared tracking}
%\title{Data hallucination for end-to-end thermal infrared tracking}
%\title{Synthetic data generation for thermal infrared tracking}
%\title{End-to-end thermal infrared trackers on synthetic data}

\author{Lichao Zhang, Abel Gonzalez-Garcia, Joost van de Weijer, Martin Danelljan, and Fahad Shahbaz Khan
\thanks{Lichao Zhang, Abel Gonzalez, Joost van de Weijer, are with the Computer Vision Center Barcelona, Edifici O, Campus UAB, 08193, Bellaterra, Spain.}
\thanks{Lichao Zhang and Joost van de Weijer are also with the Universitat Aut\`{o}noma de Barcelona, Spain.}
\thanks{Fahad Khan and Martin Danelljan are with the Link\"oping University, Sweden}}
%\author{,~\IEEEmembership{Member,~IEEE,}        ,~\IEEEmembership{,~OSA,}and,~\IEEEmembership{,~IEEE}% <-this % stops a space

%\thanks{M. Shell was with the Department of Electrical and Computer Engineering, Georgia Institute of Technology, Atlanta, GA, 30332 USA e-mail: (see http://www.michaelshell.org/contact.html).}% <-this % stops a space

\markboth{submitted to IEEE TRANSACTIONS ON IMAGE PROCESSING}
{Shell \MakeLowercase{\textit{et al.}}: Bare Demo of IEEEtran.cls for IEEE Journals}

\maketitle

\begin{abstract}
The usage of both off-the-shelf and end-to-end trained deep networks have significantly improved performance of visual tracking on RGB videos. However, the lack of large labeled datasets hampers the usage of convolutional neural networks for tracking in thermal infrared (TIR) images. Therefore, most state of the art methods on tracking for TIR data are still based on handcrafted features. 

To address this problem, we propose to use image-to-image translation models. These models allow us to translate the abundantly available labeled RGB data to synthetic TIR data. We explore both the usage of paired and unpaired image translation models for this purpose. These methods provide us with a large labeled dataset of synthetic TIR sequences, on which we can train end-to-end optimal features for tracking. To the best of our knowledge we are the first to train end-to-end features for TIR tracking.
We perform extensive experiments on VOT-TIR2017 dataset. We show that a network trained on a large dataset of synthetic TIR data obtains better performance than one trained on the available real TIR data. Combining both data sources leads to further improvement. In addition, when we combine the network with motion features we outperform the state of the art with a relative gain of over 10\%, clearly showing the efficiency of using synthetic data to train end-to-end TIR trackers.
\end{abstract}


\IEEEpeerreviewmaketitle

\section{Introduction}

Visual tracking aims to estimate a trajectory of an object through a video based on only one bounding box annotation at the beginning of the sequence. Tracking is important for applications in surveillance~\cite{emami2012role}, video understanding~\cite{renoust2016visual} and robotics~\cite{liu2012hand}.  One of the main challenges of tracking is the limited data problem: the tracker should be able to track an object based on only a single annotated bounding box. The success of a tracker is therefore very dependent on the quality of the discriminative features which are used by the tracker. 

Recently, specialized tracking subproblems have emerged. Among these is the field of tracking in thermal infrared (TIR) images, whose importance is further increasing due to improvements of thermal infrared sensors in resolution and quality~\cite{7406435,felsberg2016thermal,Kristan_2017_ICCV}. The advantage of thermal images is that they are not influenced by the illumination variations and shadows, and objects can be distinguished from the background as the background is normally colder. In addition, thermal infrared tracking can be used in total darkness, where visual cameras have no signal. Considering these advantages, thermal infrared tracking has a wide range of applications in car and pedestrian surveillance systems as well as various defense systems~\cite{gade2014thermal}. 

In recent years, Discriminative Correlation Filter (DCF) based methods~\cite{bolme2010visual, henriques2015high, danelljan2017eco} have shown to provide excellent tracking performance on existing benchmarks~\cite{wu2015object,VOT_TPAMI,mueller2016benchmark}. The DCF based trackers learn a correlation filter from example patches to discriminate between the target and background appearance. Further, the DCF based framework efficiently utilizes all
spatial shifts of the training samples by exploiting the properties of circular correlation to train and apply a discriminative classifier in a sliding
window fashion. Lately, the DCF based framework has been significantly advanced by employing high-dimensional visual features~\cite{henriques2015high,danelljan2014adaptive,danelljan2016adaptive}, powerful learning methods~\cite{song2017crest,danelljan2017eco}, reducing boundary effects~\cite{danelljan2015learning}, and accurate scale estimation~\cite{danelljan2014accurate}. Due to their superior performance in RGB tracking, some of these methods have also been applied with success to TIR~\cite{danelljan2017eco,danelljan2015learning}.

%Furthermore, the filters can be computed efficiently in the Fourier domain~\cite{henriques2015high} and the method can easily incorporate online updating~\cite{henriques2015high,danelljan2014adaptive,danelljan2016adaptive}. The DCF based methods have dominated the important tracking benchmarks~\cite{wu2015object,VOT_TPAMI,mueller2016benchmark} and have been extended with color features~\cite{van2009learning,danelljan2014adaptive}, and multi-scale~\cite{danelljan2014accurate}. They have also been applied with success to TIR~\cite{danelljan2017eco,danelljan2015learning}. Their performance, however, is very dependent on the feature representation that is used~\cite{danelljan2015learning,7406435}. 

% Paragraph on deep learning for tracking
Recently, deep learning has revolutionized the field of computer vision significantly advancing the state-of-the-art in many applications~\cite{krizhevsky2012imagenet}. Generally the deep networks are trained on a large amount of labeled training data. Despite its astounding success, the impact of deep learning on generic visual tracking (RGB data) has been limited. One of the key issues when employing deep features for tracking is the unavailability of large-scale labeled tracking data for training. Further, the tracking model is desired to be learned using a single labeled frame. Therefore, most existing deep learning based DCF trackers~\cite{ma2015hierarchical,danelljan2016beyond,danelljan2017eco} employ deep features pre-trained on the ImageNet dataset~\cite{russakovsky2015imagenet} for image classification task. Other approaches~\cite{valmadre2017end,song2017crest} have investigated the integration of DCF in a deep network by adapting the end-to-end philosophy, but did not result in major improvements over features from pre-trained networks.

%The extension of deep learning to visual tracking (for RGB data) was relatively late, and initially focused on using pre-trained deep features for tracking~\cite{danelljan2015convolutional,danelljan2016beyond,danelljan2017eco}. This is partially caused by the fact that the backpropagation for the most successful tracking approach, namely DCF, is far from evident~\cite{danelljan2015convolutional,danelljan2017eco,song2017crest}. And even when applying backpropagation, the reported results were only slightly higher than those obtained with pre-trained networks~\cite{gundogdu2017good}. As consequence off-the-shelf deep features are often for visual tracking in RGB videos. 

% paragraph on the small datasets for FIR, and the usage of handcrafted features for FIR tracking
Even more than for RGB tracking, introducing deep learning to TIR tracking is hampered by the absence of large datasets. The datasets which are available for thermal infrared videos are relatively small. Moreover, there is no ImageNet counterpart of infrared still images on which a large network could be pre-trained. Therefore, the usage of handcrafted features remains dominant for TIR tracking. For instance, the top three trackers in VOT-TIR2017~\cite{Kristan_2017_ICCV,VOT_TPAMI} are still exploiting handcrafted features. The winner~\cite{yu2017dense} of VOT-TIR2017 challenge employs HOG~\cite{dalal2005histograms} and motion features. Further, the other top-performing methods ~\cite{zhu2016beyond,7406435} are based on handcrafted features. The success of these methods show that handcrafted features are still the best choice for TIR tracking. 

%As a consequence of the absence of large datasets to train end-to-end discriminative features for TIR tracking, handcrafted feature are still the current best choice for this modality. 

%The winner of VOT-TIR2017 challenge is DSLT~\cite{yu2017dense} whose Estimated Average Overlap (EAO) is 0.399. They use HOG~\cite{dalal2005histograms} and motion features. Also EBT~\cite{zhu2016beyond} and SRDCFir~\cite{7406435} are based on handcrafted features. As a consequence of the absence of large datasets to train end-to-end discriminative features for TIR tracking, handcrafted feature are still the current best choice for this modality.

%EBT achieves 0.368 by using edges features. Through the introduction of motion features, SRDCFir obtains a jump result from 0.243 to 0.357. Currently, the best public performance on VOT-TIR2017~\cite{Kristan_2017_ICCV} is 0.401 by DSLT in~\cite{yu2017dense}.\alertJW{Do they use deep features?} \lichao{None of them use deep features}
%\alertJW{WHAT IS THE CURRENT STATEOFTHEART ??}.

% paragraph on image to image translation models (pix2pix, cycle consistency)
Deep learning has also resulted in fast progress in generative models which are able to generate samples from complex image distributions~\cite{goodfellow2014generative}. These models have been further extended to image-to-image translation models~\cite{isola2017image} which allow to learn mappings between image domains. A further extension of this work allows to learn mappings between unpaired domains~\cite{zhu2017unpaired}, which is based on the observation that transferring an image to another domain and then transferring it back to the first domain should result in the same image which was provided as input. One of the more interesting applications of these generative networks is that they can be used to construct synthetic datasets of small data domains, such as TIR. In this work, we show that labeled data from RGB can be translated to TIR data, and the labels can be transferred. 

% paragraph on main idea paper.
%\minisection{Contribution:} In this work we tackle the key limited-data problem for TIR-tracking by utilizing recent developments in image-to-image translation methods ~\cite{isola2017image,zhu2017unpaired}. The idea is to automatically transfer labeled RGB tracking videos to the TIR domain. We can also automatically transfer the labels from the RGB videos to these synthetic TIR videos. The resulting data can then be used to extract discriminative deep features for the TIR domain. The advantage is that we can generate the TIR counterpart of the available RGB tracking datasets which are much larger compared to the current TIR tracking datasets. We will show that a tracker trained on only synthetic data can outperform trackers trained on available labeled TIR data (see Fig.~\ref{fig:qualResECO}). We show the effectiveness of our synthetic TIR data generation model on the large dataset KAIST~\cite{hwang2013multispectral}. Furthermore, extensive evaluations on the latest TIR tracking challenges~\cite{Kristan_2017_ICCV} verify the efficiency of our different models trained on these generated TIR datasets. Finally, we show state of the art results when our TIR features are combined with motion features on the TIR tracking challenges.

\begin{figure*}
    \centering
    \includegraphics[width=\textwidth]{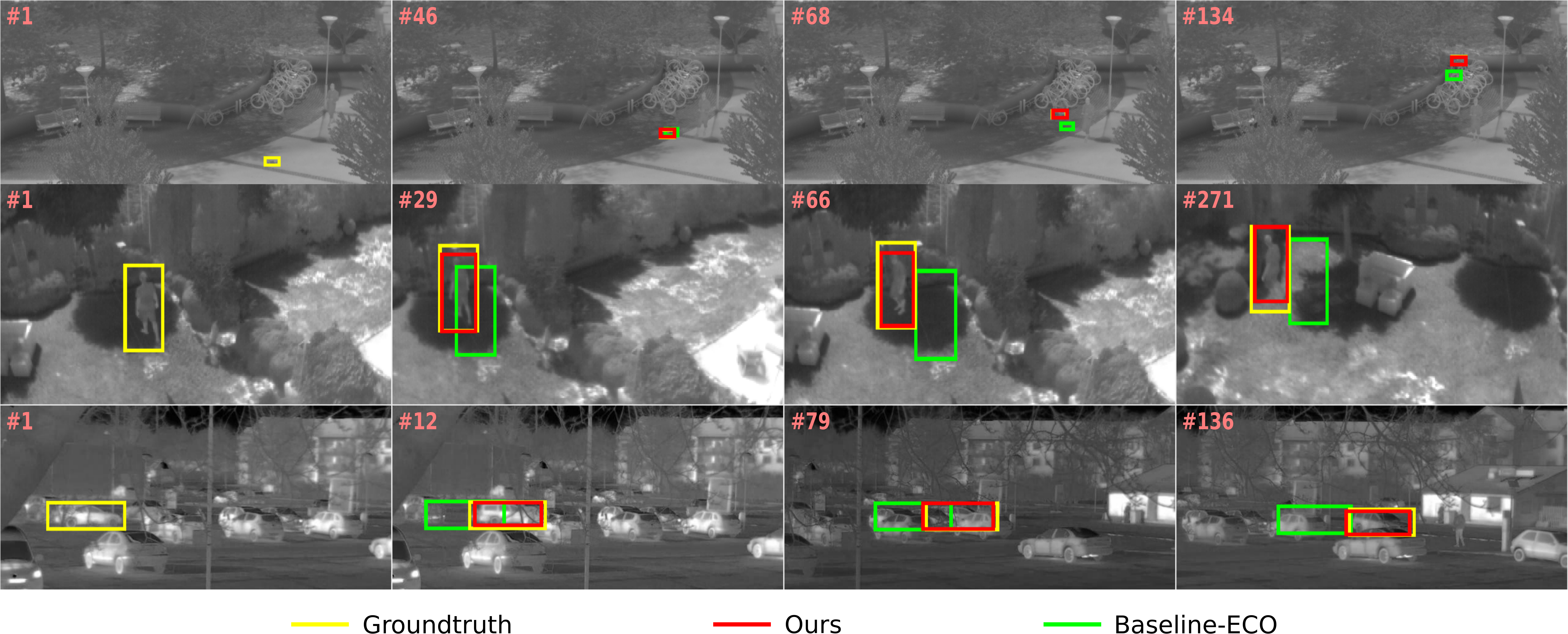}
    \caption{Qualitative comparison of our approach trained on generated data only (red) with baseline ECO~\cite{danelljan2017eco} (green) on the \textit{quadrocopter2}, \textit{garden} and \textit{car2} videos. The ground truth bounding box is provided in yellow. Owing to the synthetic TIR data our model is able to follow the object successfully in case of out-of-plane rotation, partial occlusion and scale changes.}
    \label{fig:qualResECO}
\end{figure*}

In this work we tackle the key limited-data problem for TIR-tracking by utilizing recent developments in image-to-image translation methods ~\cite{isola2017image,zhu2017unpaired}.
The idea is to automatically transfer RGB tracking videos to the TIR domain. We can then automatically transfer the labels from these RGB videos to the synthetic TIR videos. The resulting data can then be used to extract discriminative deep features for the TIR domain. The advantage is that we can generate the TIR counterpart of the available RGB tracking datasets which are much larger compared to the current TIR tracking datasets. {The main contributions of the paper are: 
\begin{itemize}
\item We address the scarcity of labeled data for TIR tracking. Therefore, we propose a framework which transfers RGB data to synthetic TIR data. The labels available for the RGB data are also transfered to the TIR data, resulting in a large synthetic TIR data set for tracking.
\item We are the first to perform end-to-end training for TIR tracking, showing that this can significantly improve results (see Table~\ref{table:models}). We also show that a tracker trained on only synthetic data can outperforms trackers trained on available labeled TIR data (see Fig.~\ref{fig:qualResECO}).
%\item  We show the effectiveness of our synthetic TIR data generation model on the large dataset KAIST~\cite{hwang2013multispectral}. 
\item We perform extensive evaluations on the latest TIR tracking challenge~\cite{Kristan_2017_ICCV} verifying the efficiency of our different models trained on synthetic TIR datasets. We show that when combined with motion features our method obtains state of the art on the TIR tracking challenge.
\end{itemize}}

% overview paper
The rest of the paper is organized as follows. In section~\ref{sec:related} we briefly discuss related work. In section~\ref{sec:corrfilter} we introduce the standard correlation filter and the current end-to-end deep correlation filter. In section~\ref{sec:gan} we describe the prevalent generative adversarial networks and present our generated synthetic tracking videos. In section~\ref{sec:exps} we present our experiments on standard thermal infrared tracking dataset. In section~\ref{sec:conclude} we conclude our work and plan our further research.
%In RGB dataset, the most outstanding results are definitely obtained by using the CNN features.

% figure 1

%%%%%%%%%%%%%%%%%%%%%%%%%%%%%%%%%
%%%%%%%%%%%%%%%%%%%%%%%%%%%%%%%%%

\section{Related Work}\label{sec:related}

\subsection{DCF tracking}
% Added by Fahad
In recent years, discriminative correlation filter (DCF) based tracking methods have shown excellent performance in terms of accuracy and robustness on benchmark tracking datasets~\cite{VOT_TPAMI,wu2015object}. The DCF based trackers aim at learning a correlation filter in an online fashion from example image patches to discriminate between the target and background appearance. The seminal work of~\cite{bolme2010visual} was restricted to a single feature channel (grayscale image). Later, the DCF framework was extended to use multi-dimensional handcrafted features by~\cite{galoogahi2013multi,henriques2015high,danelljan2014adaptive}, such as HOG~\cite{dalal2005histograms} and Color Names~\cite{van2009learning}. Some of the recent advances in DCF frameworks can be attributed to reducing boundary effects~\cite{danelljan2015learning}, robust scale estimation~\cite{danelljan2014accurate}, integrating context~\cite{cf_ca_tracking}, and adding a long-term memory component~\cite{ma2015long}.

Even after more than five years of flourishing, discriminative correlation filter based tracking is still the mainstream in single object tracking. Recent modifications on DCF include: Mueller et al.~\cite{cf_ca_tracking} sample four context patches around the target and incorporate these to regularize the regression function, which has the same effect as hard negative mining. Alan et al.~\cite{lukezic2017discriminative} enlarge the search region and improve tracking of non-rectangular objects by using spatial maps to restrict the correlation filter. They also give the learned filter adaptive channel-wise weights, which improves the quality of the filter. Kiani et al.~\cite{kiani2017learning} use a mask to crop the object in the spatial domain and get a new closed-form solution of the correlation filter in the Fourier domain by embedding the mask matrix into the formulation. This yields significantly more shifted examples unaffected by boundary effects.

Compared to handcrafted features (e.g. HOG~\cite{dalal2005histograms}, intensity and Color Names~\cite{van2009learning}), deep CNN features significantly improve the robustness of the tracker against geometric variations, resulting in a significant improvement of the performance~\cite{danelljan2015convolutional}. This mainly caused by the high discrimination of deep features, since CNNs are trained on the large dataset ILSVRC2012~\cite{russakovsky2015imagenet}. Later, Ma et al.~\cite{ma2015hierarchical} propose to encode the target appearance on several convolutional layers and each layer has a corresponding correlation filter. This hierarchical architecture locates targets by maximizing the response of each layer with different weights. They find an optimized position in a coarse to fine way. Directly using different layers may not take full advantage of the CNN features because of the discrete distribution of features. To exploit the continuity between different layers of networks, Danelljan et al.~\cite{danelljan2016beyond} propose to learn a convolution operator in the continuous spatial domain called CCOT. As CCOT is very slow, Danelljan et al.~\cite{danelljan2017eco} propose to factorize the convolution operator to reduce the dimensions of feature maps. Then they use GMM to generate samples, which significantly accelerate the tracker, enabling the tracker to run in real-time, while still maintaining the same or higher accuracy.
%Nowadays, correlation filters have evolved as the basic patch matching method for tracking.

\subsection{TIR tracking}
Currently, the leading TIR trackers still employ handcrafted features in their models. Yu et al.~\cite{yu2017online} propose structural learning on dense samples around the object. Their tracker uses edge and HOG features and transfers them into the Fourier domain, to obtain a real-time tracker. Later they extend this work, called DSLT~\cite{yu2017dense}, by integrating HOG~\cite{dalal2005histograms} and motion features. With this tracker they won the VOT-TIR2017 challenge~\cite{Kristan_2017_ICCV}.
%attributes to its effective feature representation for infrared objects. we introduce an effective feature representation for infrared objects. 1D motion feature 
%
Another TIR tracker, called EBT~\cite{zhu2016beyond}, uses edge features to devise an objectness measure specific for each instance. This enables the generation of high quality object proposals and the use of richer features.
Concretely, for each proposal they extract a 2640-dimensional histogram feature as well as a 5-level pyramid computed from the intensity channel. They achieve the runner-up position in the VOT-TIR2017 challenge.
%\AG{careful! EBT uses edge features to get better proposals, but the actual classification of the proposals uses histogram features in RGB and intensity}
% to extends the field range of proposals generation to the whole image by instance-specific objectness measure and suppresses distractors by focusing the object model on hard proposals, and by d, it performs well on 
SRDCFir~\cite{7406435} extends the SRDCF~\cite{danelljan2015learning} tracker for TIR data by adding motion features. SRDCF is a DCF-based tracker that introduced a spatial regularization function to penalize those DCF coefficients that reside outside the target region, which mitigates the damaging boundary effects present in the traditional DCF.

Another branch for TIR tracking combines the input TIR data with the visual modality, concretely with the image intensity given as a grayscale image. For example, Li et al.~\cite{li2016learning} propose an adaptive fusion scheme to incorporate information from grayscale images and TIR videos during tracking.
Similarly, the approach in~\cite{li2017grayscale} samples a set of patches around the object and extracts a joint sparse representation in both grayscale and TIR modalities.

The usage of generating other modalities was pioneered by Hoffman et al.~\cite{hoffman2016learning}. They used generation of depth data to improve classification on the abundant-data modality (RGB), whereas we use data generation as a source of labeled data for the scarce-data modality (TIR). Xu et al.~\cite{xu2017learning} use a network which generates TIR images to pre-train the weights. These weights are then applied in a network which is used on RGB data with the aim to improve tracking of pedestrians. Other than them, we use the generation of TIR data for data augmentation; we create large synthetic labeled data sets of TIR data to be able to train end-to-end features for TIR data. 

\subsection{Adversarial image-to-image translation}
% \AG{not a contribution, one paragraph should be enough I guess}
Generative Adversarial Networks (GANs)~\cite{goodfellow2014generative} have achieved promising results in several tasks such as image generation~\cite{denton2015deep}, image editing~\cite{perarnau2016invertible}, and representation learning~\cite{salimans2016improved}. 
The conditional variants of GANs~\cite{mirza2014conditional} enable to condition the image generation on a selected input variable, for example, an input image. 
In this case, the task becomes image-to-image translation, and this is the variant we use here.
The general method of Isola et al.~\cite{isola2017image}, pix2pix, was the first GAN-based image-to-image translation work that was not designed for a specific task (e.g. colorization~\cite{zhang2016colorful}).
The architecture is based on an encoder-decoder with skip connections~\cite{ronneberger2015u} and it is trained using a combination of two losses: a conditional adversarial loss~\cite{goodfellow2014generative} and a loss that maps the generated image close to the corresponding target image.
This method achieves excellent results, but requires matching pairs of training images, which limits the applicability of the model as such data might not be easily accessible.
In order to overcome this limitation, Zhu et al.~\cite{zhu2017unpaired} extended this model to the case in which paired data is not available.
Their method, called CycleGAN, relies on the assumption that mapping an image from the input domain to the target and then back to the input (i.e. the cycle) should result in the identity function.
Based on this, they add a cycle consistency loss that enforces the correct reconstruction of the input image resulting of the composition of the two mappings. 
They demonstrate the effectiveness of their method for multiple tasks such as edges to real images or photo enhancement.
In this paper, we use image-to-image translation to generate a synthetic large-scale TIR tracking dataset from a labeled RGB dataset, with the goal of learning better deep features for tracking. 

%the  same  idea  for  conditional image generation applications, such as future prediction in videos~\cite{mathieu2015deep,vondrick2016generating}, image inpainting~\cite{pathak2016context}. 

%  Inspired by the concept that there are many image-to-image translation problems in computer vision, namely mapping an image in one domain to a corresponding image in another domain. 
% For instance, super-resolution can be considered as a problem of mapping a low-resolution image to a corresponding high-resolution image; colorization can be considered as a problem of mapping a gray-scale image to a corresponding color image.
% So considering the limits of future prediction and application of the image generation in image-to-image translation for different domain. 
% We adopt now current prevalent generated adversarial network to transform the existed video frames to a special domain which is lack of corresponding images. Comparatively in this way large video datasets matching the aim domain can be generated.

%%%%%%%%%%%%%%%%%%%%%%%%%%%%%%%%%
%%%%%%%%%%%%%%%%%%%%%%%%%%%%%%%%%

\section{Method overview}
We aim to train end-to-end deep features for tracking in TIR data. However,  to train effective deep features for TIR data, we need a large dataset of labeled TIR videos. Unfortunately, the amount of labeled TIR data is very scarce.
To the best of our knowledge, only BU-TIV dataset~\cite{wu2014thermal} contains a considerable amount of labeled TIR videos, but most of them depict only one object class (pedestrian). Therefore, most state of the art TIR tracking methods are still based on hand-crafted features~\cite{yu2017dense,zhu2016beyond,7406435}.
On the other hand, there are vast amounts of RGB videos labeled for tracking~\cite{wu2015object,VOT_TPAMI}. One solution could therefore be to use the pre-trained features which are optimal for tracking in RGB data for TIR data. However, this is unlikely to be optimal because TIR and RGB data differ significantly. 

\begin{figure}
    \centering
    \includegraphics[width=0.5\textwidth]{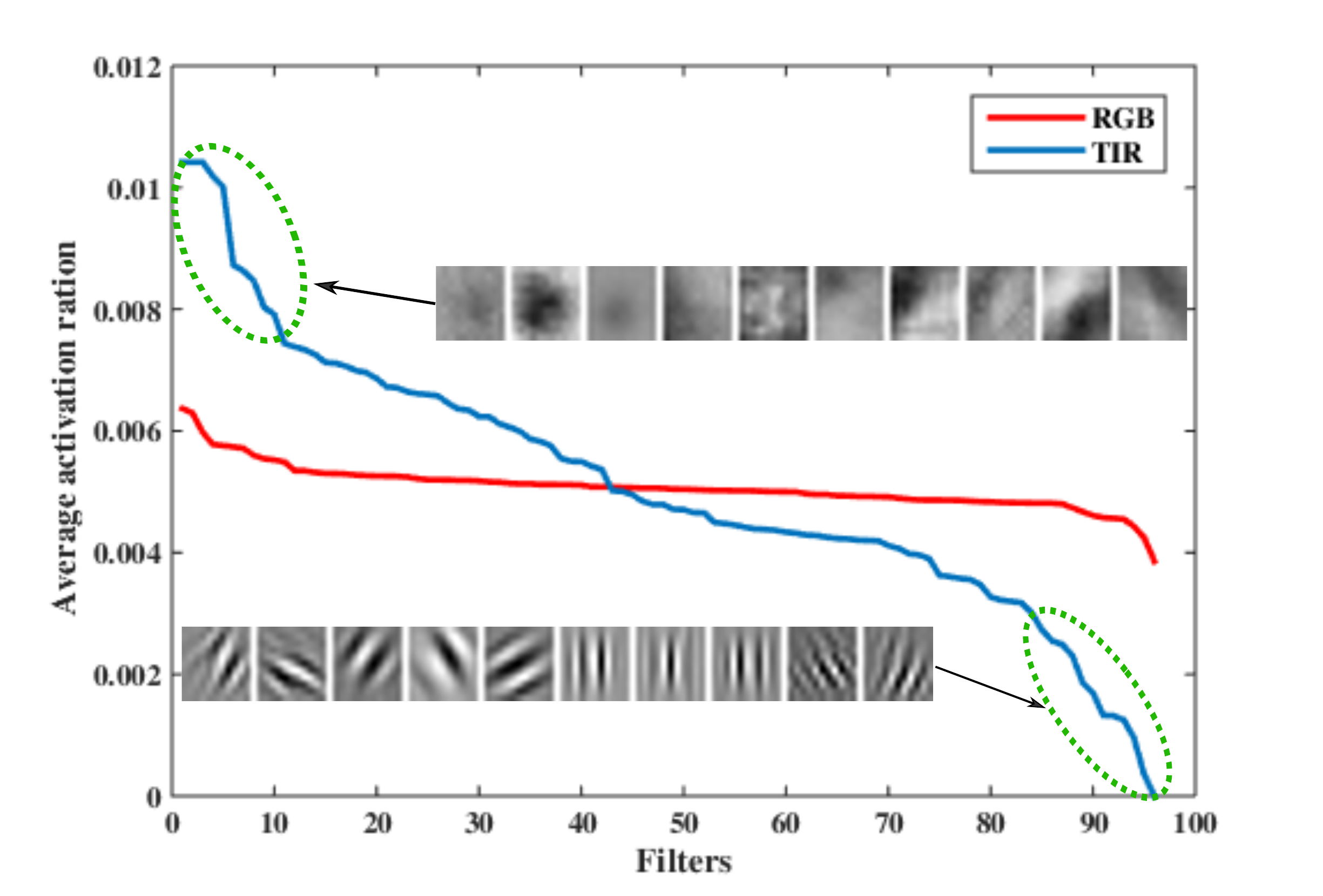}
	\caption{{Average activation of filters from the first layer of pre-trained AlexNet~\cite{krizhevsky2012imagenet} on the test set of KAIST~\cite{hwang2013multispectral} for RGB and TIR images.}}
    \label{fig:filter}
\end{figure}

{To illustrate the difference in nature of RGB and TIR data we measure the average activation of the 96 filters of the first layer of a pre-trained AlexNet on the KAIST dataset. This dataset contains both RGB and TIR images of the same scenes (a similar study for depth images has been performed in~\cite{song2017depth}). The pre-trained network is trained to recognize objects in RGB images (i.e. on ImageNet). In Fig.~\ref{fig:filter} we show the results. The graph shows the average activation of the filters in descending order. When applied to data which is similar to that on which the network is trained, the average activations tend to produce a uniform distribution. 
This can be seen for RGB images where most filters yield the same average activation and only a few filters differ from this pattern. When we perform the same experiment on TIR data the pattern changes. We can now observe clear differences between filters which have a higher average activation and filters which have lower average activation. This shows that these filters are probably not optimal for TIR tracking. When we look at the exact filters which have low and high activation, we see that low frequency patterns (blobs and edges) are prevalent for the TIR data, whereas high-frequency filters are seldom in TIR data. This is not surprising since most textures, responsible for most of the high-frequency content of images, do not appear in TIR data. In conclusion, given the different nature of the image statistics of RGB data and TIR data it is probable that a network which is trained on TIR data would outperform a network trained on RGB data. }

{In this paper, we aim to address the problem of data scarcity of labeled videos for tracking in TIR data. We do this by exploiting the vast amount of labelled RGB videos which are available, in combination with recent advances in image-to-image translation techniques. We will use these image-to-image translation models to transfer large labeled RGB datasets to synthetic TIR dataset together with the tracking annotations. As a result we now have a large labeled synthetic TIR dataset.  We use this synthetic TIR dataset to train end-to-end deep networks to obtain optimal TIR features for tracking.
Then we plug the optimal TIR feature model into a state-of-the-art tracker. Here we use ECO~\cite{danelljan2017eco}. An overview of our method is provided in Fig.~\ref{fig:pipeline}. In the following section we detail the various parts of our algorithm.}

%Compared with baseline RGB data training model, our synthetic TIR data %training model can significantly improve the performance of tracker on VOT-TIR2017 dataset~\cite{Kristan_2017_ICCV}, results are shown in Table~\ref{table:models}.

%As in Fig.~\ref{fig:pipeline}, we propose to use the efficient image-to-image translation model to transfer large label-available RGB dataset to synthetic TIR dataset. And use these synthetic TIR dataset to end-to-end train the feature networks and obtain the optimal TIR feature model for tracking. Then we plug the optimal TIR feature model into a state-of-the-art tracker, here we use ECO~\cite{danelljan2017eco}. Compared with baseline RGB data training model, our synthetic TIR data training model can significantly improve the performance of tracker on VOT-TIR2017 dataset~\cite{Kristan_2017_ICCV}, results are shown in Table~\ref{table:models}.
%Therefore, we consider several methods to translate labeled RGB dataset to synthetic TIR datasets, to be able to train end-to-end network which are optimized for tracking in TIR data.

\section{Deep Learning Features for Correlation Filter Tracking}
\label{sec:corrfilter}
In this section, we introduce the standard correlation filter and the current end-to-end deep correlation filter. Then we describe Efficient Convolution Operators (ECO)~\cite{danelljan2017eco} which we will use as the correlation filter for our experiments. 

\subsection{Correlation Filter Tracking}
The conventional discriminative correlation filters (DCF) formulation~{\cite{henriques2015high}}, learns a linear correlation filter $f$ that discriminates the target appearance from the background. The target location is predicted by applying the correlation filter to a sample feature map. The desired filter $f$ can be obtained by minimizing a least squares objective:
\begin{align} 
 %E(f) = \sum^M_{j=1} \|S_{f}\{x_j\}-y_j  \|^2+\lambda \sum^D_{d=1} \|f^d\|^2.
 E(f) = \left\|\sum^D_{d=1} f^d*x^d-y  \right\|^2+\lambda \sum^D_{d=1} \|f^d\|^2. 
 \label{eq:cf} 
 \end{align} 
 %\mathbb{E}\left(f\right) = \left \|\sum^D_{d=1} f^d*J^l(x)-y_j \right \|^2+\lambda \sum^D_{l=1} \left \|f^l \right \|^2,
Here $*$ denotes circular correlation. $x^d$ denotes feature maps of training samples $x$, where the layer $d\in\{1,\dots,D\}$. $f^d$ denotes channel $d$ of filter $f$. $y$ is the regression target and $\lambda$ is a regularization weight to control over-fitting. A closed-form solution is obtained in the Fourier domain,
%\martin{This is not true for the multi-sample case. Eq. (2) needs to be changed to a single sample by removing the sum (i.e. M=1) in that case.}
\begin{eqnarray}
%\hat{f}^d = \frac{\hat{\phi}_d\{x\} \hat{y}^*}{\sum^D_{d=1}\hat{\phi}_d\{x\} (\hat{\phi}_d\{x\})^* + \lambda}\,.
\hat{f}^d = \frac{\hat{x}^d \hat{y}^*}{\sum^D_{d=1}\hat{x}^d (\hat{x}^d)^* + \lambda}.
\label{eq:filter}
\end{eqnarray} 
Where the $\hat{y}^*$ denotes the complex conjugate of the discrete Fourier transform $\mathcal{F}(y)$.
% \begin{eqnarray}
%  \hat{f}^d = \frac{\hat{J}^l(x) \odot \hat{g}^*}{\sum^d_{k=1}\hat{J}^l(x)\odot (\hat{J}^l(x))^* + \lambda},
%  \label{eq:filter}
%  \end{eqnarray} 
% \subsection{End-to-end training of DCF Trackers}

%\fkmd{Add transition from ECO to simple DCF used in CFnet for end-to-end learning.}

%Despite of the fact the deep features successful are plugged into correlation filter tracker, the deep networks are normally a pre-trained trained on the ILSVRC~\cite{russakovsky2015imagenet} dataset trained in a loss function for solving classification task. 

\begin{figure*}
    \centering
    \includegraphics[width=\textwidth]{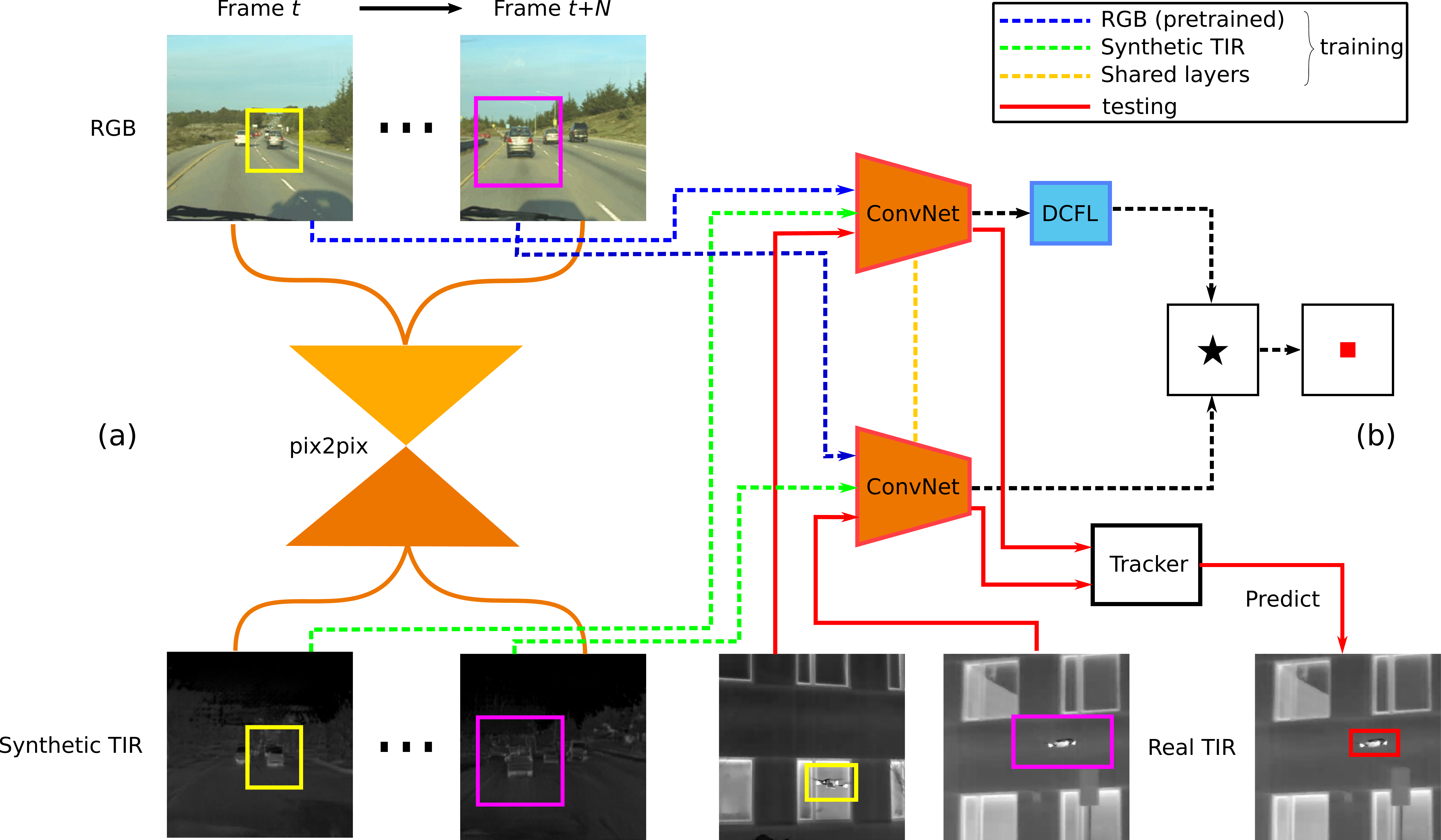}
    \caption{ {Overview of our approach. (a) Image-to-image translation component (proposed in~\cite{isola2017image}) for generating a large labeled synthetic TIR tracking dataset. We use blue dashed line to represent the baseline RGB training model and the green dashed line represents our proposed synthetic data training model. After the translation of RGB data to TIR data, we acquire enough suitable data for end-to-end training networks for TIR tracking. (b) Two-branch architecture for training the network to obtain adaptive features for TIR tracking (proposed in~\cite{valmadre2017end}). The optimal correlation filter is computed in the discriminative correlation filter layer (DCFL) for the image processed in the upper branch. This filter is then applied on the image in the bottom branch. }} 
    %The yellow dashed line between networks indicates that they share their weights.
  %  Overview of our approach. On the left is the image to image translation component for generating a large labeled synthetic TIR tracking dataset. On the right is the two-branch architecture for training the network to obtain adaptive features for TIR tracking. The optimal correlation filter is computed in the discriminative correlation filter layer (DCFL) for the image processed in the upper branch. This filter is then applied on the image in the bottom branch. The yellow dashed line between networks indicates that they share their weights.}
    \label{fig:pipeline}
\end{figure*}

Recently, researchers have proposed several methods for end-to-end training of features for tracking: CFNet~\cite{valmadre2017end}, DCFNet~\cite{wang2017dcfnet}, and CFCF~\cite{gundogdu2017good}. All use a two-branch Siamese network, of which one branch is used to compute the optimal correlation filter, which is applied in the other branch to obtain the response map (see Fig.~\ref{fig:pipeline}). Both branches share the weights of their convolutional layers. Training is performed with paired images from same video. It backpropagates the gradients through the discriminative correlation filter layer (DCFL) with a closed-from solution~\cite{valmadre2017end}. Surprisingly, trackers based on end-to-end training only slightly outperformed off-the-shelf features. It should be noted that all end-to-end trackers train on RGB datasets, mainly on the ImageNet Large Scale Visual Recognition Challenge (ILSVRC15)~\cite{russakovsky2015imagenet}, and no results for end-to-end tracking on other modalities like TIR are available.  

In this paper, we use the end-to-end CFNet training procedure proposed by Bertinetto et al.~\cite{valmadre2017end}. This method obtains stable and fast network training due to their Fourier domain implementation of the discriminative correlation filter layer. Other than them we will apply it to TIR tracking. Since current available datasets for TIR tracking are rather small, we propose in the next section our approach to generating synthetic TIR tracking data from labeled RGB tracking data. 

\subsection{Efficient Convolution Operators}

Previously, we have explained how to train end-to-end features for tracking. These features can be used in different discriminative correlation filter methods. In our work we use the Efficient Convolution Operator (ECO) \cite{danelljan2017eco} method, shown to obtain state-of-the-art results while being computationally efficient. However, its original implementation is based on features extracted from a pre-trained CNN model trained on the ImageNet 2012 classification dataset~\cite{russakovsky2015imagenet}. Even though these features are extracted from a model which is trained for image classification, ECO obtains excellent results for tracking. In our experiments we combine ECO with the end-to-end trained features for TIR tracking.

The ECO tracker aims at combining shallow and deep features by learning a multi-channel continuous convolution filter in a joint optimization scheme across all feature channels. Furthermore, it learns a projection matrix, to reduce the dimensionality of high-dimensional features. Here we briefly describe the training and inference procedures applied in the ECO tracker.

%We base our approach on the recently introduced Efficient Convolution Operator (ECO) \cite{danelljan2017eco} method, shown to obtain state-of-the-art results while being computationally efficient. The ECO tracker aims at combining shallow and deep features by learning a multi-channel continuous convolution filter in a joint optimization scheme across all feature channels. Furthermore, it learns a projection matrix, to reduce the dimensionality of high-dimensional features. Here we briefly describe the training and inference procedures applied in the ECO tracker.

ECO learns the target model parameters based on a set of training samples {$\{x_j\}_1^M$} and corresponding labels {$\{y_j\}_1^M$}. {The label function $y_j$ consists of the desired target scores at all spatial locations in the corresponding training sample $x_j$. It is defined as a periodically repeated Gaussian function centered at the sample location~\cite{danelljan2016beyond}.}
% We denote the set of training samples as  and each training sample $x_j$ is associated with a labeled detection score function $y_j$, which is a periodically repeated Gaussian function
Each training sample contains multiple feature layers $x_j^d\in\mathbb{R}^{N_d\times N_d}$, where $N_d$ is the spatial resolution of layer $d\in\{1,\dots,D\}$. These feature layers correspond to both shallow and deep features of varying resolutions. The tracker predicts the target location using the target score operator, defined as
\begin{align}
  \label{eco_score}
  S_{f,P}\{x\} = \sum_{d=1}^Df^d*PJ_d\{x^d\} \,.
\end{align}
Here, $x$ is the input sample and $f$ is the learned filter that predicts the detection score function $S_{f,P}\{x\}$ of the target. The sample $x$ is first interpolated to the continuous domain using the operator $J_d${, by applying a cubic spline kernel in the Fourier domain (see~\cite{danelljan2016beyond} for details)}. The projection matrix $P$ is then applied to reduce the dimensionality of the feature space.

The detection score operator is learnt via minimization of a least squares objective,
\begin{align}
  \label{eco_loss}
  E(f) = &\sum_{j=1}^M\alpha_j\|S_{f,P}\{x_j\}-y_j\|^2 + \sum_{d=1}^D\|wf^d\|^2 + \lambda\|P\|_F^2 \,.
\end{align}
Here, the projection and filter are regularized by {a constant $\lambda$}. The spatial regularization weight function $w$ is employed to mitigate the effects of periodic repetition \cite{danelljan2015learning}. Each sample $x_j$ is weighted by $\alpha_j$, based on a learning rate parameter $\gamma$. The label functions $\{y_j\}_1^M$ are set to Gaussian functions centered at the target location.
%$\lambda^{\mathrm{proj}}$
% \begin{align}
%   \label{label}
%   y(t_1,t_2) = \expo{-\frac{(t_1-u_1)^2}{2(a\sigma)^2}}\expo{-\frac{(t_2-u_2)^2}{2(b\sigma)^2}},
% \end{align}
% periodically repeated. Here $(u_1,u_2)$ is the estimated target position. The variance of the Gaussian is controlled by the target size $(a,b)$ and a factor $\sigma$. 

Using Parseval's formula an equivalent loss is obtained as,
\begin{align}
  \label{eco_loss_fourier}
  E(f) =& \sum_{j=1}^M\alpha_k\|\widehat{S_{f,P}\{x_j\}}-\hat{y}_j\|^2 \!+ \!\sum_{d=1}^D\|\hat{w}*\hat{f}^d\|^2 + \lambda \|P\|_F^2.
\end{align}
Here $\hat{\cdot}$ denotes the Fourier coefficients. {We learn the projection matrix $P$ jointly with the filter $f$ in the first frame by applying Gauss-Newton and adopting the Conjugate Gradient method~\cite{nocedal2006numerical} for each iteration.
In subsequent frames, the resulting normal equations are efficiently solved using the method of Conjugate Gradients, assuming a fixed $P$.} For more details, we refer to \cite{danelljan2017eco,danelljan2016beyond}.

\section{Generating TIR images}\label{sec:gan}
%\AG{I'm not a big fan of the format of these subsubsections, can we change them? For example, remove the colon?}
%\AG{I really haven't used the keyword `hallucination' here because to me it sounds like something that is also done at test time. For a strictly training time data augmentation procedure I think generate/synthesize/translate are more adequate.}
%\alertJW{We could consider a pix2pix figure of our approach}
%% Intro, why do we need this

{In this section we discuss image-to-image translation methods and compare them for the task of transferring RGB to synthetic TIR data.}

\subsection{Image-to-image translation methods}
%% Methods overview
We use two different image-to-image translation methods to transform labeled RGB videos into labeled TIR videos. 
First, we use pix2pix~\cite{isola2017image}, which requires paired training data. 
Therefore, we need matching frames in both RGB and TIR, which we can obtain from multispectral video datasets such as KAIST~\cite{hwang2013multispectral}. 
% The training pairs exist of an RGB frame and a matching TIR frame. 
Second, we use CycleGAN~\cite{zhu2017unpaired}, an extension on pix2pix that can be trained from unpaired data. 
As a consequence, any videos in the RGB and TIR modalities can be used to train CycleGAN.
{Despite the higher availability of unpaired training data, we expect the weaker supervision of CycleGAN to result in synthesized TIR images of lower quality. In this section, we present both translation methods and experimentally confirm this intuition. In later sections, we generate TIR data using only pix2pix given its empirically superior performance.}

% Brief description common to both methods
Both methods are based on Generative Adversarial Networks (GANs)~\cite{goodfellow2014generative} conditioned on input images.
GANs consist of two networks, generator $G$ and discriminator $D$, that compete against each other. 
The generator tries to generate samples that resemble the original data distribution, whereas the discriminator tries to detect whether samples are real or have been generated by $G$.
When the GAN architecture is conditioned on an input image, the task becomes image-to-image translation.
In our case, the input image is a color frame from an RGB video and the target is the matching frame in the TIR modality. 

% Here we show that this approach is effective on a wider range of problems. 
% We transfer the videos in visual light spectral to Thermal InfraRed spectral(TIR), considering now there are few datasets for thermal infrared tracking which only exists in the VOT-TIR~\cite{felsberg2016thermal} dataset. 
% Don't mention no more TIR data for training networks. 
% We investigate the effect of different generated datasets. 
% We use the pix2pix generated model and cycle generated model to transform the visual tracking dataset to TIR tracking dataset.

\subsubsection{Paired - pix2pix}
% As conditional adversarial network not only learn the mapping from input image to output image, but also learn a loss function to train this mapping. This makes it possible to apply the same generic approach to problems that traditionally would require very different loss formulations. 
% Pix2pix~\cite{isola2017image} demonstrates that this approach is effective at synthesizing photos from label maps, reconstructing objects from edge maps, colorizing images, transferring thermal to color and among other task.

pix2pix~\cite{isola2017image} is an effective, task-agnostic method that can be applied to translate between many domain pairs, including maps to satellite pictures, edge maps to real pictures, or grayscale images to color images.   
The generator is based on an encoder-decoder architecture with skip connections (U-Net~\cite{ronneberger2015u}).
The discriminator is a convolutional PatchGAN~\cite{li2016precomputed}, which classifies each local image patch independently, making it especially suited for modifying textures or styles.
  
Let $x$ be an image from the input domain $X$ and $y$ an image from the target domain $Y$. 
In pix2pix, both the generator and discriminator are conditioned on the input image $x$.
The conditional GAN objective function is defined as
\begin{eqnarray}
\mathcal{L}_{cGAN}\left(G,D\right) &=& \mathbb{E}_{x,y} [\log D(x,y)] \nonumber \\ 
	&+& \mathbb{E}_{x,z}[ \log\left(1-D\left(x,G\left(x,z\right)\right)\right)],
\label{eq:cGAN}
\end{eqnarray} 
where $z$ is a random noise vector used as input for the generator.
Additionally, pix2pix also includes an L1 loss to increase the sharpness of the output images 
\begin{eqnarray}
\mathcal{L}_{L1}\left(G\right) = \mathbb{E}_{x,y,z} [\left \| y-G(x,z)  \right \|_1].
\label{eq:L1_con}
\end{eqnarray} 
The final objective function is the weighted sum of these two losses.
Following the original adversarial training~\cite{goodfellow2014generative}, $G$ tries to minimize this final objective while $D$ tries to maximize it:
\begin{eqnarray}
G^*  = \mathop {\arg \min\limits_G } \mathop {\max }\limits_D \mathcal{L}_{cGAN}\left(G,D\right)+\lambda\mathcal{L}_{L1}\left(G\right).
\label{eq:con_obj}
\end{eqnarray}
We translate an RGB video to TIR by applying pix2pix independently for each video frame.
The original model of~\cite{isola2017image} achieves mild stochasticity in its outputs by keeping the dropout layers at test time, which are normally used only during training.
In our case, this is not only unnecessary but also damaging, as it makes the output video less stable.
For this reason, we only use dropout layers during training.

\subsubsection{Unpaired - CycleGAN} 

Paired data might be hard to come by for particular tasks including RGB to TIR conversion, as the amount of paired videos in both modalities is rather limited.
Zhu et al.~\cite{zhu2017unpaired} present CycleGAN, a method for learning to translate between image domains when paired examples are not available.
The main idea consists in adding a cycle consistency loss, based on the assumption that mapping an image $x\in X$ to domain $Y$ and back to $X$ should leave it unaltered.
{
For this reason, besides the classic generator $G:X\rightarrow Y$, CycleGAN also learns a generator to perform the inverse mapping $F:Y\rightarrow X$.
The method is then trained with a weighted combination of an unconditional adversarial loss~\cite{goodfellow2014generative} 
and the cycle consistency loss in both directions
\begin{eqnarray}
\mathcal{L}_{cyc}\left(G,F\right) &=& \mathbb{E}_{x} [\left \| F(G(x))-x  \right \|_1] \nonumber \\ 
	&+&\mathbb{E}_{y} [\left \| G(F(y))-y  \right \|_1],
\label{eq:Lcyc2}
\end{eqnarray}
For more details, please see~\cite{zhu2017unpaired}. As in the pix2pix model, we apply CycleGAN independently per frame, and we remove the dropout layers at test time to generate a more stable video output.
}

\begin{figure*}[t]
\centering
     \includegraphics[width=\textwidth]{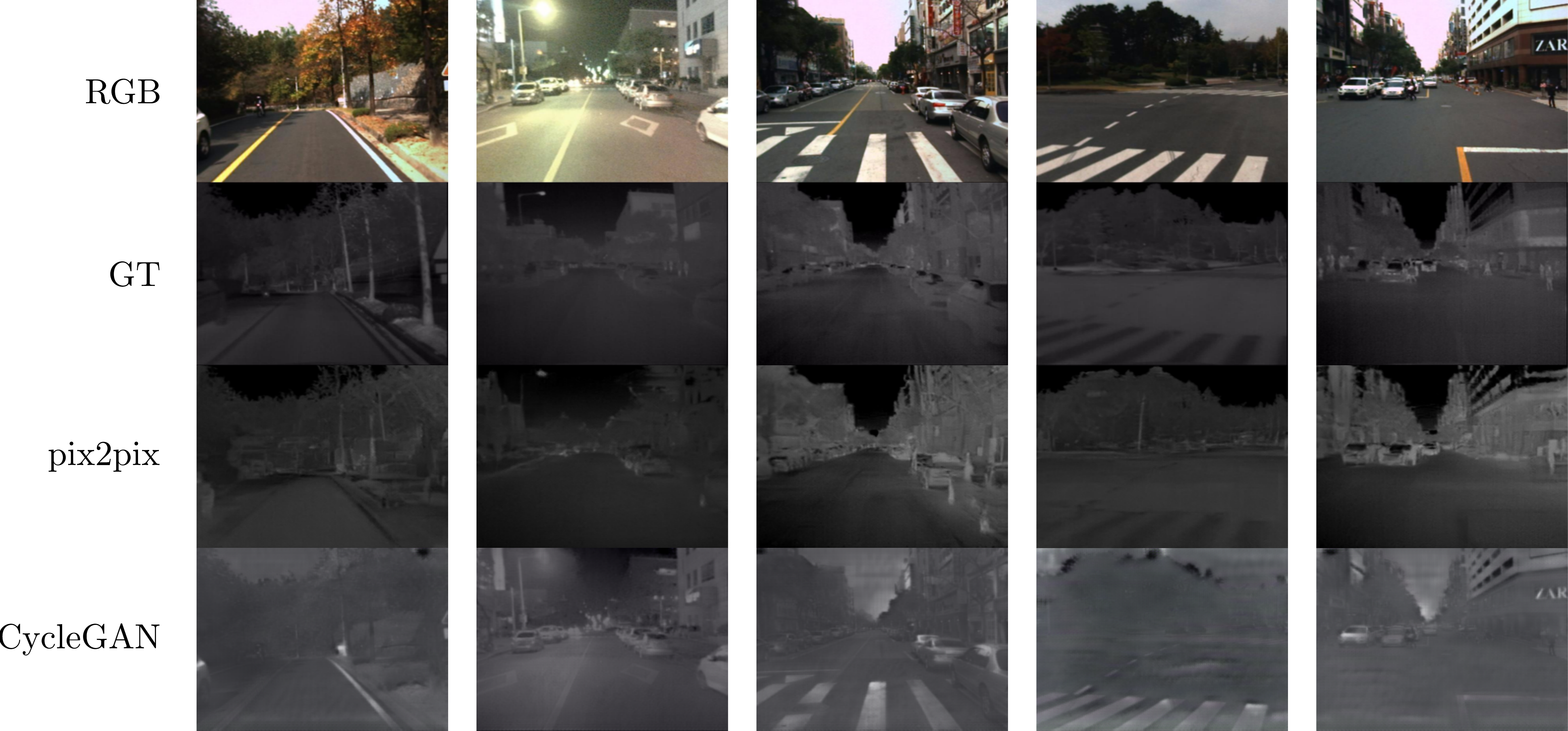}
      \caption{Results for the two image translation methods considered: pix2pix and CycleGAN. The video frames are taken from the test set of KAIST~\cite{hwang2013multispectral}, and have not been seen during training.}
      \label{fig:GANresults}
\end{figure*}

\subsection{Datasets}
\setlength{\tabcolsep}{4pt}
\begin{table}
{
    \begin{center}        
        \begin{tabular}{c|c|c|c}
            \hline%\noalign{\smallskip}
            \multirow{2}{*}{Type} & \multirow{2}{*}{Dataset}
      & \multicolumn{2}{c}{Number of images} \\ \cline{3-4}
      & & RGB & TIR \\ \hline
      \multirow{6}{*}{Paired} & KAIST~\cite{hwang2013multispectral} & 50,184 & 50,184 \\
      & CVC-14~\cite{gonzalez2016pedestrian} & 8,473 & 8,473 \\
      & OSU Color Thermal~\cite{davis2005two} & 8,545 & 8,545 \\
      & VAP Trimodal~\cite{palmero2016multi} & 5,924 & 5,924 \\
      & Bilodeau~\cite{bilodeau2014thermal} & 7,821 & 7,821 \\
      & LITIV2012~\cite{torabi2012iterative} & 6,325 & 6,325\\ \cline{2-4}
      & total & 87,088 & 87,088 \\ \hline
      \multirow{6}{*}{Unpaired} & VOT2016~\cite{kristan2016VOT} & 21,455 & - \\
      & VOT2017~\cite{Kristan_2017_ICCV} & 4,049 & - \\
      & OTB~\cite{wu2015object} & 58610 & -\\
      & ASL~\cite{portmann2014people} & - & 6,490 \\
      & Long-term~\cite{gade2013long} & - & 47,423 \\
      & InfAR~\cite{gao2016infar} & - & 46,121 \\ \cline{2-4}
      & total & 84,114 & 100,034 \\ \hline
        \end{tabular}
        \caption{{Datasets used for training the image-to-image translation models. We test all models using a subset of three videos from the official test set of KAIST~\cite{hwang2013multispectral}.}}
        \label{table:datasetsGAN}
    \end{center}
}
\end{table}
%
% We train our image translation models with a combination of multiple datasets.
%
{
We consider multiple datasets for training our image translation methods, spanning the two presented supervision levels: paired and unpaired.  
Table~\ref{table:datasetsGAN} details the number of images of all the considered datasets.
Among the paired datasets, the biggest and most relevant is} KAIST Multispectral Pedestrian Dataset~\cite{hwang2013multispectral}, which contains a significant amount of aligned images in the RGB and TIR modalities, captured from a moving vehicle in different urban environments and under different lighting conditions.
We follow the official data split~\cite{hwang2013multispectral} as in~\cite{isola2017image} and use all the frames from training videos for training.
We evaluate both image translation methods using 3 randomly left out videos from the test set, {amounting to 5,728 images.
Train and test sets have no videos in common.
}

{
Other paired datasets include images of people captured under different conditions: pedestrians during day or night (CVC-14~\cite{gonzalez2016pedestrian}), static cameras at a busy intersection (OSU Color-Thermal Database~\cite{davis2005two}) or in different positions and zooms (LITIV2012 dataset~\cite{torabi2012iterative}, interactions in indoor scenes with controlled lighting settings (VAP Trimodal People Segmentation Dataset~\cite{palmero2016multi}), or moving in different planes (Bilodeau et al.~\cite{bilodeau2014thermal}).
This amounts to a total of 87K image pairs. }

{
We use all paired datasets to train both pix2pix and CycleGAN.
Additionally, we collect an RGB-TIR unpaired dataset as extra training data for CycleGAN. 
As RGB data we include} all the sequences from VOT2016~\cite{kristan2016VOT}, VOT2017~\cite{Kristan_2017_ICCV}, and OTB~\cite{wu2015object}. 
As TIR data we include the TIR images from ASL~\cite{portmann2014people}, Long-term~\cite{gade2013long}, and InfAR~\cite{gao2016infar}.
This amounts to a total of about 230K images, almost $5\times$ more images than the paired training dataset. 

\subsection{Implementation details}
We train all networks in pix2pix and CycleGAN from scratch, initializing the weights from a Gaussian distribution with zero mean and standard deviation of 0.02. 
We use the same network architectures as in the original papers~\cite{isola2017image,zhu2017unpaired}. 
As in~\cite{isola2017image}, we apply random jittering by slightly enlarging the input image and then randomly cropping back to the original size.
We train pix2pix for 10 epochs, with batch size 4 and learning rate 0.0002.
CycleGAN is trained for 3 epochs, batch size 2 and learning rate 0.0002. 
Note that both models are trained for an equivalent number of iterations given the size of their training sets. 

\subsection{TIR image translation quality}
In order to test the two image translation methods considered we select a random subset of the test set of KAIST~\cite{hwang2013multispectral}, amounting to about 10\% of the entire dataset.
We translate the RGB videos into TIR using pix2pix or CycleGAN, and then compute the Euclidean distance of the translations with the TIR ground-truth images. 
Finally, we average the distance for all frames.
pix2pix obtains an average distance of 35.3, whereas CycleGAN obtains 69.5.
This demonstrates the superiority of pix2pix for this task, showing how a paired training signal is more valuable than the unpaired counterpart, despite the bigger training dataset of the latter.
Fig.~\ref{fig:GANresults} shows a qualitative comparison of both approaches.
We can observe how the translated images using pix2pix are clearly superior to those translated by CycleGAN.
Moreover, they look remarkably similar to the ground-truth TIR images, confirming the validity of the proposed data augmentation approach.
Therefore, we select pix2pix as our method to generate TIR tracking data from RGB videos.

\begin{figure}
    \centering
    \includegraphics[width=0.5\textwidth]{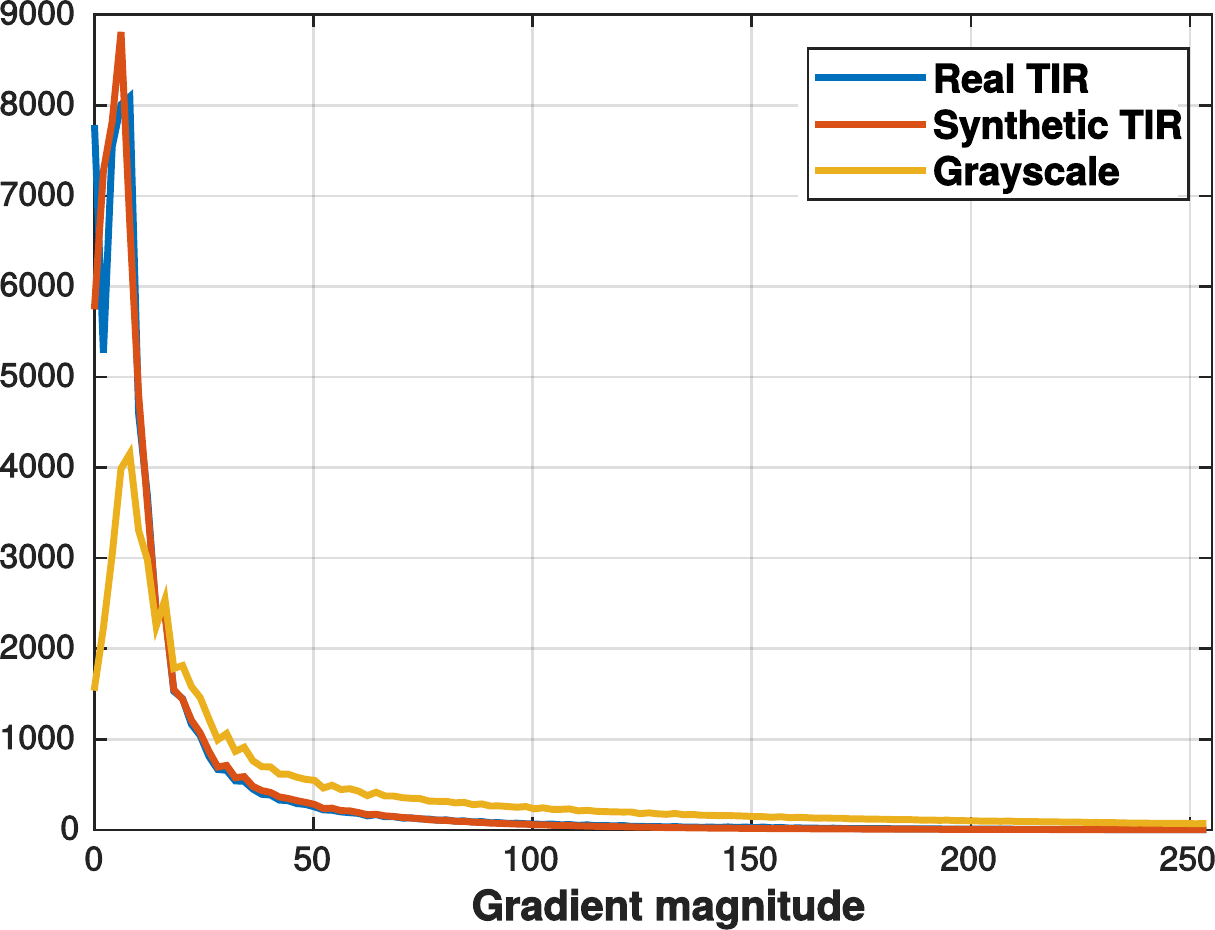}
	\caption{{Histogram of the gradient magnitude for real and synthetic TIR data computed on the test set of KAIST~\cite{hwang2013multispectral}. For comparison we have also added the gradient magnitude histogram for grayscale images from which the synthetic dataset has been generated.}}
    \label{fig:gradients}
\end{figure}

{In addition we compare the statistics of the image gradients of real TIR data and synthetic TIR data generated by pix2pix. The histogram of the gradient magnitude for both datasets on the test set of KAIST is provided in Fig.~\ref{fig:gradients}. We have also added the gradient magnitude of the grayscale images from which the synthetic dataset is generated. The results show that the gradient magnitude of the synthetic data closely follows that of the real data. Only small variations can be seen for low magnitude gradients. The similarity of the image statistics of real and synthetic data suggests that trackers trained on the synthetic data could be successful on real TIR data.}

\section{Experimental Results}\label{sec:exps}
% 0. reconstruction error, conclusion pix2pix (moves to section 4).

%EXP1
%1a. compare handcraft, pretrained, with fine-tuned on real and generated on pix2pix
%1b. compare two methods of combining real and generated data
%EXP 2
%2a. compare various results with motion masks
%2b. compare SoA.

%VAP~\cite{palmero2016multi} same as Litiv-VAP

\subsection{Datasets}

We train several versions of our tracker using both real and generated TIR tracking data, {summarized in Table~\ref{table:datasetsTracker}. 
}
As real TIR data we use BU-TIV~\cite{wu2014thermal}, ASL~\cite{portmann2014people}, and OTCBVS~\cite{davis2005two}. 
The predominant class in these datasets is human/pedestrian, although BU-TIV~\cite{wu2014thermal} includes some vehicles and ASL~\cite{portmann2014people} also contains animals like cat and horse. 
We select all those sequences that include annotated bounding boxes around the objects, {leading to a total of 375K bounding boxes from 34K images.}
% There is a total of 32,635 frames from 18 different videos.
% Most videos contain several objects, which leads to a total of 375K bounding boxes from tracked objects. 
On the other hand, we generate synthetic TIR tracking data using the RGB videos from VOT2016~\cite{kristan2016VOT}, VOT2017~\cite{Kristan_2017_ICCV}, and OTB~\cite{wu2015object}, which are standard tracking benchmarks used by the community.
% VOT2016 consists of 60 videos with a total of 21455 frames.
% VOT2017 replaces the 10 easiest videos of VOT2016 with 10 new videos, adding 4049 new frames.
% Finally, OTB has 98 videos that amount to 58,610 frames overall.
In total, we obtain 168 TIR videos with tracking annotations by translating the original RGB frames using pix2pix and transferring the corresponding bounding box annotations.
{The total number of bounding boxes is 4.5$\times$ greater than in the real TIR images.}
Furthermore, the generated TIR videos contain a wider variety of object classes than the real TIR videos. This increases the generality of the learned deep features. 
In both cases, we leave out around $10\%$ of the videos during training as validation set.
% validation of GAN, vot2016 6 videos, 2426 frames; vot2017 1 video (ants1), 325 frames; OTB 9 videos, 5226 frames. (training: 152 videos, 76137 frames; validation: 16 videos, 7977 frames)

\label{sec:datasets}
\setlength{\tabcolsep}{4pt}
\begin{table}
{
    \begin{center}        
        \begin{tabular}{c|c|c|c|c}
            \hline%\noalign{\smallskip}
            Type & Dataset & Videos & Images & Bounding-boxes\\ \hline
      \multirow{4}{*}{Real} & BU-TIV~\cite{wu2014thermal} & 5 & 23,393 & 34,7291 \\ 
      & ASL~\cite{portmann2014people} & 13 & 6,490 & 7,773 \\ 
      & OTCBVS~\cite{davis2005two} & 4 & 4861 &	19,944 \\ \cline{2-5}
      & Total & 22 & 34,744 & 375,008\\ \hline
      \multirow{4}{*}{Generated} & VOT2016~\cite{kristan2016VOT} & 60 & 21,455 & 21,455 \\
      & VOT2017~\cite{Kristan_2017_ICCV} & 10 & 4,049 & 4,049 \\
      & OTB~\cite{wu2015object} & 98 & 58,610 & 58,610 \\ \cline{2-5}
      & Total & 168 & 84,114 & 84,114\\ \hline
        \end{tabular}
        \caption{{Datasets used for training the tracker, using real TIR data or generated TIR data from RGB images.}}
        \label{table:datasetsTracker}
    \end{center}
}
\end{table}
We evaluate our TIR tracker on the VOT-TIR2017 dataset~\cite{Kristan_2017_ICCV}, which is identical to VOT-TIR2016 dataset~\cite{felsberg2016thermal} as the 2016 edition of this benchmark was far from being saturated.
It contains 25 TIR videos of varying image resolution, with an average sequence length of 740 frames adding up to a total of 13,863 frames.
Each sequence has been manually annotated with exactly one bounding box per frame around a particular object instance.
There is a wide variety of object classes, including pedestrian, animals such as rhino or bird, and vehicles like quadrocopter or car. 
Moreover, the dataset includes extra annotations in the form of attributes, either at frame level (e.g. camera motion, occlusion) or at the sequence level (e.g. blur, background clutter).
{This test dataset has no videos in common with the RGB modality of VOT2016-17 used for training.}

\subsection{Evaluation measures and protocol}
We follow the measures and evaluation protocol proposed by the VOT-TIR2017 benchmark~\cite{felsberg2016thermal}.
The two primary measures are accuracy (A) and robustness (R), which have been shown to be highly interpretable and only weakly correlated~\cite{vcehovin2016visual}.
Accuracy is computed as the overlap (intersection over union) between the predicted track region and the ground-truth bounding box, averaged over frames.
The VOT protocol establishes that when the evaluated tracker fails, i.e. when the overlap is below a given threshold, it is re-initialized in the correct location five frames after the failure.
In order to reduce the positive bias introduced by this protocol, the accuracy measure ignores the first ten frames after the re-initialization when computing the average overlap. 
Robustness measures the number of times the tracker fails for each sequence and then takes the average over all sequences.
These two measures are conflated into a third, the Expected Average Overlap (EAO), which is the main measure used to rank the trackers.
The EAO estimates the expected average overlap of a tracker for a particular sequence of a fixed, short length.
We refer the reader to~\cite{VOT_TPAMI} for more details.

Besides the standard VOT metrics, we also report results following the One-Pass Evaluation (OPE) protocol originally proposed in~\cite{wu2015object}.
The most standard evaluation metric used with this protocol is success rate.
For each frame in the test video, we compute the overlap between the predicted track and the ground-truth bounding box.
A predicted track is considered successful if its overlap with the ground-truth is above a particular threshold. 
We obtain a success plot by evaluating the success rate at different overlap thresholds. 
Conventionally, the Area Under the Curve (AUC) of the success plot is reported as a summary measure. 
Note how this protocol does not reset the tracker in case of failure.
% The precision plot can be acquired in a similar way, but usually the representative precision at the threshold of 20 pixels is reported. 
We use the VOT toolkit~\cite{Kristan_2017_ICCV} to compute the measure and plot the results. 

\subsection{Implementation details}
% Before we train our model, we also collect some real thermal infrared dataset are from BU-TIV~\cite{wu2014thermal}, ASL~\cite{portmann2014people} and OTCBVS~\cite{davis2005two}. We select the sequences with annotations. They are totally 18 videos, 32,635 frames and 61549 objects for detection. 

%seq2, seq2, seq2, lab1-green and lab1-red from BU-TIV. They are totally 5 videos, 23393 frames and 347291 objects for detection.

We train CFNet following~\cite{valmadre2017end}. 
{We perform tests with three different networks as base model: AlexNet~\cite{krizhevsky2012imagenet}, VGG-M~\cite{chatfield2014return}, and ResNet-50~\cite{he2016deep}.
}
As in~\cite{valmadre2017end}, we reduce the total stride of the networks from 16 to 4 by changing the stride of the first and second pooling layers from 2 to 1 in AlexNet, and that of second convolutional and pooling layers in VGG-M.
This allows us to obtain bigger feature maps, which benefits the correlation filters.
For fairness, we apply this modification to all trained models.
%For tracking the object spatial information should be preserved as much as possible, 
As training input data for the network, we randomly pick object regions from pairs of images from the same video.
Specifically we crop a centered region on the object of approximately twice the object's size, and resize it to $125\times 125$ pixels.
We use Stochastic Gradient Descent (SGD) with momentum of 0.9 and weight decay of 0.0005 to fine-tune the network, which is pre-trained for image classification on ILSVCR12~\cite{russakovsky2015imagenet}.
The learning rate is decreased logarithmically at each epoch from $10^{-4}$ to $10^{-5}$. 
The model is trained for 50 epochs with mini-batches of size 128.

For the baseline tracker ECO~\cite{danelljan2017eco} {(Fig.~\ref{fig:pipeline}, blue dashed lines)}, we use the recommended settings (`OTB\_DEEP\_settings') detailed in the code provided by the authors~\cite{ecocode}.
ECO is an RGB tracker, so we have adapted the following parameters given the different nature of TIR data.
{Following~\cite{nam2016learning,park2018meta}} we use the feature map of the third convolutional layer as the input of the correlation filter, {(convolutional block in case of ResNet-50).
We confirm these results in the following section.}
We reduce the learning rate used to update the correlation filter from the 0.009 used for RGB data to 0.003.
A smaller learning rate is more suitable for TIR data, as TIR images have less detailed information than RGB, for example lacking texture, and thus the object appearance remains more stable during tracking.
In order to optimally leverage the learned CNN features, we do not add the dimensionality reduction step at the output of each layer as in~\cite{danelljan2017eco}. 
ECO uses this to increase the tracker's efficiency, which is not a priority in our work.

Upon acceptance we will make the different trained models available for the community.

% As we want to see the CNN features' abilities, we only utilize the CNN as the tracker's feature and don't compress dimensionality of each output layer. 

% \setlength{\tabcolsep}{4pt}
% \begin{table}
%     \begin{center}        
%         \begin{tabular}{c|ccc|ccc}
%             \hline%\noalign{\smallskip}
%             \multirow{2}{*}{Tracker} 
%       & \multicolumn{3}{c|}{without motion features} 
%           & \multicolumn{3}{c}{with motion features} \\ %\hline
%  		& EAO 			& A 			& R & EAO &A	&R\\ \hline 
% handcrafted 	& 0.235			&0.60 			&2.74&0.361 &0.62 &1.12 \\ %\hline
% pretrained		& \0.28	&\textbf{0.65}	&2.36&0.390 &0.65	&1.06 \\ 
% real 			& 0.289 		&0.64			&2.34&0.410 &0.63	&1.07 \\
% generated 		& 0.300			&0.64 			&2.23&0.419&0.64	&0.99 \\ \hline
% generated $\rightarrow$ real & 0.307 &0.64 	&2.06  &0.429  		& 0.65&\textbf{0.79}\\ 
% generated + real 	& \textbf{0.325} &\textbf{0.65} &\textbf{1.97}&\textbf{0.448}&\textbf{0.67}	&0.89 \\   \hline        
%         \end{tabular}
%         \caption{Comparison of different tracker variants with and without adding motion features. Results are on the VOT-TIR2017 benchmark~\cite{Kristan_2017_ICCV} with ResNet-50~\cite{he2016deep} as base network. Boldface indicates the best results. In both cases, the best results are achieved when combining both real and generated TIR data. \AG{Lichao: update table with ResNet values}}
%         \label{table:models}
%     \end{center}
% \end{table}

\begin{figure}
    \centering
    \includegraphics[width=0.5\textwidth]{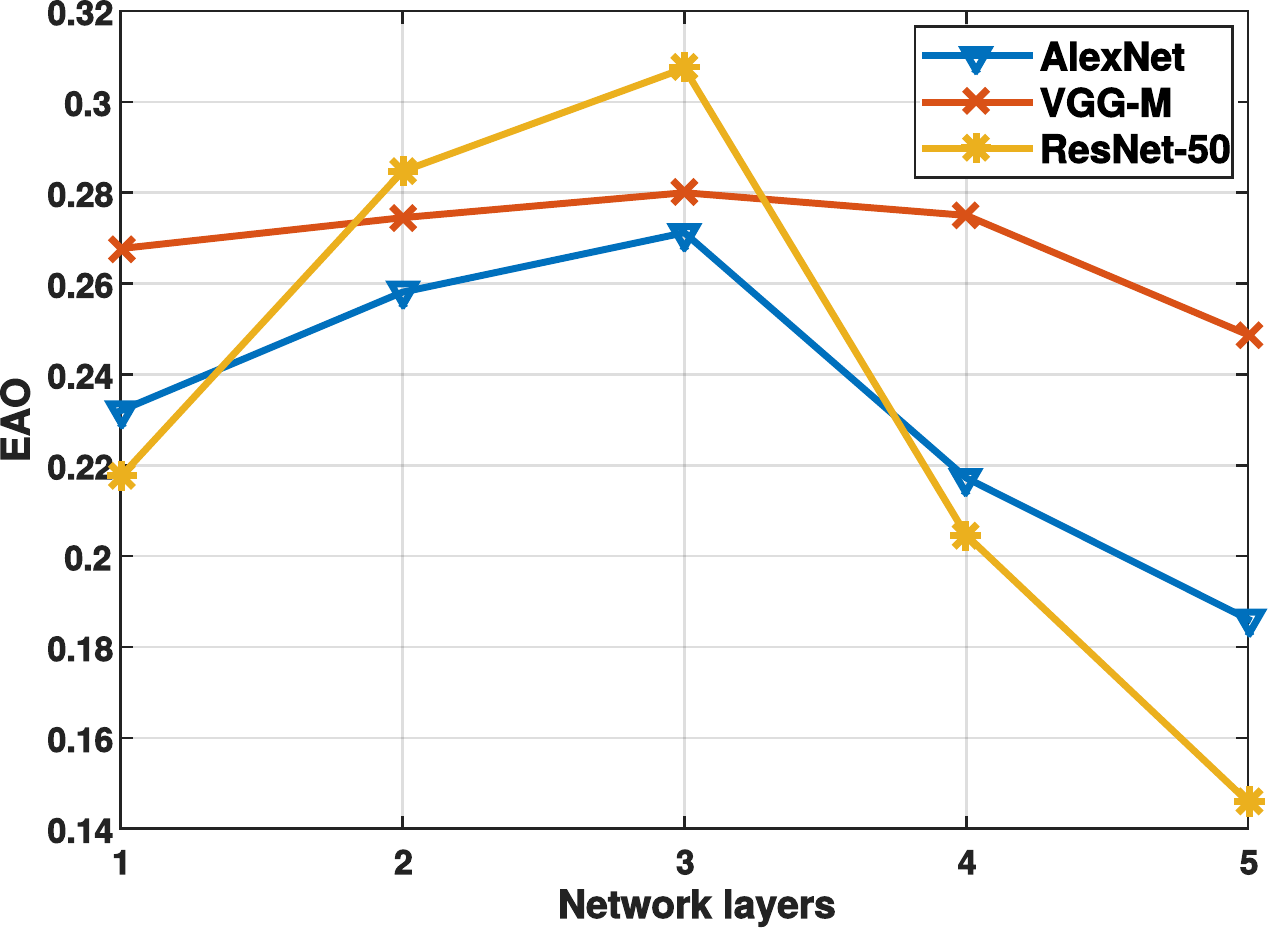}
	\caption{{The EAO on VOT-TIR2017~\cite{Kristan_2017_ICCV} when using deep features extracted from different layers.}}
    \label{fig:layers}
\end{figure}

\setlength{\tabcolsep}{4pt}
\begin{table}
    \begin{center}        
        \begin{tabular}{c|ccc|ccc}
            \hline%\noalign{\smallskip}
            \multirow{2}{*}{Tracker} 
      & \multicolumn{3}{c|}{without motion features} 
          & \multicolumn{3}{c}{with motion features} \\ %\hline
 		& EAO 			& A 			& R & EAO &A	&R\\ \hline 
handcrafted 	& 0.235			&0.60 			&2.74&0.361 &0.62 &1.12 \\ %\hline
pretrained		& {0.307} 	& 		{0.62} 	&{2.00} &{0.381}  &{\textbf{0.69}} 	&{1.06}  \\ 
real 			& {0.316}  		&{0.62} 			&{2.01} &{0.409}  &{0.67} 	&{1.24}  \\
generated 		& {0.321} 			&{\textbf{0.63}}  			&{2.00} &{0.419} &{0.65} 	&{0.83}  \\ \hline
generated $\rightarrow$ real & {0.331}  &{0.61}  	&{1.76}   & {0.429}  		& {0.63} &{0.82} \\ 
generated + real 	&{\textbf{0.347}}  &{\textbf{0.63}}  &{\textbf{1.68}} &{\textbf{0.436}} &{0.65} &{\textbf{0.80}}  \\   \hline        
        \end{tabular}
        \caption{Comparison of different tracker variants with and without adding motion features. Results are on the VOT-TIR2017 benchmark~\cite{Kristan_2017_ICCV} with ResNet-50~\cite{he2016deep} as base network. Boldface indicates the best results. In both cases, the best results are achieved when combining both real and generated TIR data. }
        \label{table:models}
    \end{center}
\end{table}
{
\subsection{Network layers}
\label{sec:layers}
Our tracker uses deep features from a particular network layer.
Previous works~\cite{nam2016learning,park2018meta} selected mid-level features from the third convolutional layer as optimal for tracking in RGB videos.
Here, we validate this choice for TIR data by analyzing the performance of the selected tracker across all layers for the three networks considered.
We perform these experiments using only pre-trained features, i.e., we do not fine-tune the networks for tracking.
Fig.~\ref{fig:layers} presents the tracking performance measured by EAO on VOT-TIR2017~\cite{Kristan_2017_ICCV} as a function of the network layer.
Trackers that use features extracted from the third layer offer the best results, and thus we select this features for the remainder of the paper.
}

\begin{figure}
    \centering
    \includegraphics[width=0.5\textwidth]{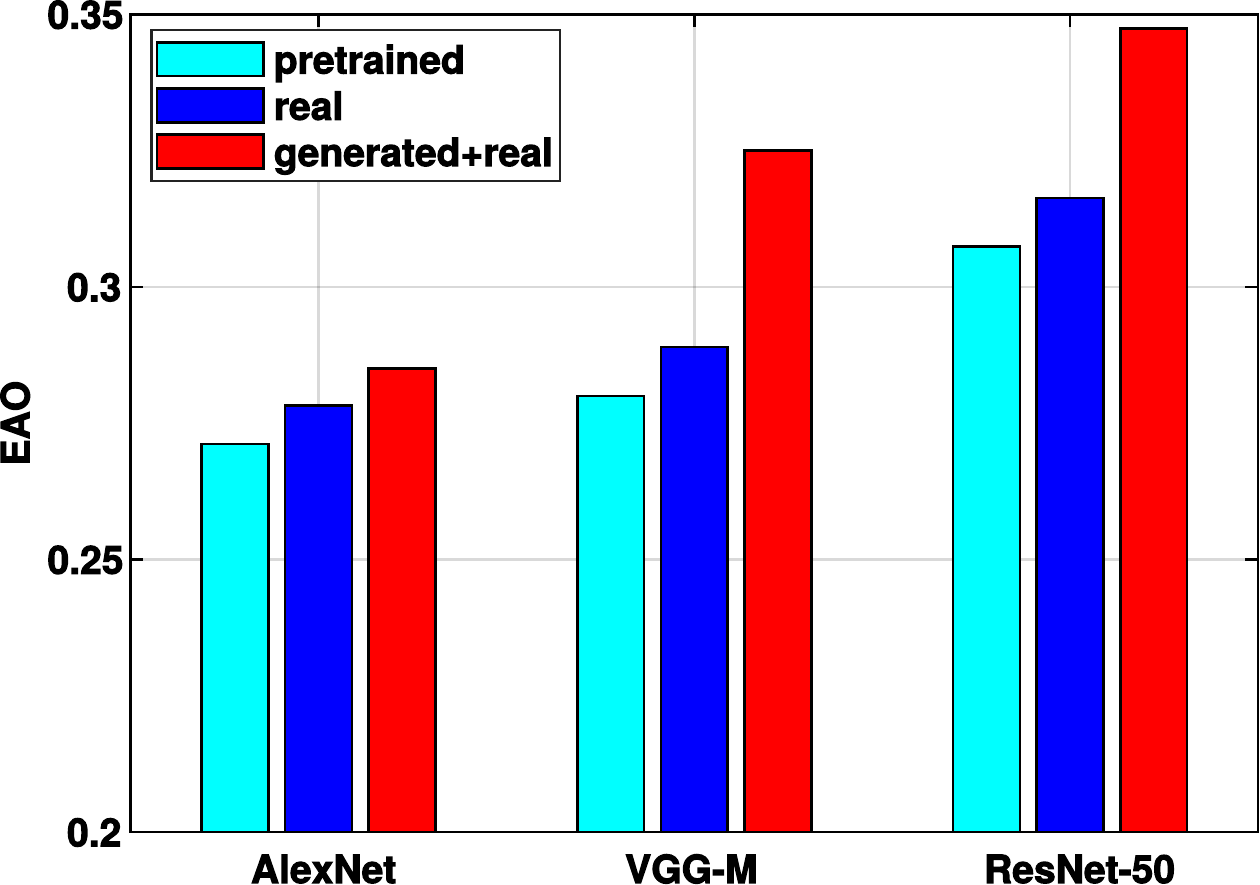}
	\caption{{The EAO on VOT-TIR2017~\cite{Kristan_2017_ICCV} when using deep features extracted from different networks. Our synthetic data can benefit general networks for fine-tuning.}}
    \label{fig:topos}
\end{figure}

{
\subsection{Network architectures}
In this section, we experiment with all three base networks with different types of data used for training.
All models use ECO~\cite{danelljan2017eco} as base tracker, in some cases with the adaptations detailed in section~\ref{sec:corrfilter}.
We consider two baselines, `pretrained' and `real'. 
The first baseline uses features from the corresponding CNN pre-trained for the image classification task.
On the other hand, `real' is also fine-tuned using real TIR tracking datasets (sec.~\ref{sec:datasets}).
Our tracker (`generated + real') combines both real TIR and synthesized from RGB with pix2pix for the fine-tuning process.
Fig.~\ref{fig:topos} presents these results.
For all base networks, fine-tuning helps when learning effective features for tracking.}
This shows that the generated data is complementary to the available real data, making the generated data beneficial even when a good amount of real data is available.
{
Moreover, the gain granted by fine-tuning the network is significantly higher when augmenting the training dataset with our generated TIR data.
The performance boost is especially remarkable for higher capacity models such as ResNet-50, since networks with more parameters require more data to train.
For all following experiments, we use ResNet-50 as base network for the trackers.
}
%
% \AG{Introduce the other two networks here and present analysis}

\subsection{Results on real and generated data}
{We now present more detailed results for different configurations of our ECO tracker with ResNet-50.}
We include another baseline (`handcrafted') that employs handcrafted features as it is prevalent in TIR tracking~\cite{yu2017dense,zhu2016beyond,7406435}, and thus has not been trained using data.
The variant called `generated' is fine-tuned using only our generated TIR data with pix2pix.
{
Finally, we consider another way of combining real and generated data to train the model, `generated $\rightarrow$ real', which besides uses a two-step fine-tuning mechanism, first using generated data and then real data.
This is opposed to `generated + real', which fine-tunes using both real and generated data simultaneously without distinction.
}
Table~\ref{table:models} presents the results for all these models using metrics EAO, A, and R on the VOT-TIR2017 dataset~\cite{Kristan_2017_ICCV}.
%for instance DSLT~\cite{yu2017dense}, EBT~\cite{zhu2016beyond} and SRDCFir~\cite{7406435}.

First, we can observe how the use of deep features is fundamental for the success of this tracker, given the low accuracy of the handcrafted model.
Simply using pre-trained features already provides a significant improvement in terms of EAO.
Fine-tuning this model on real data brings further benefits. Interestingly, only fine-tuning on the generated data using pix2pix results in better performance than fine-tuning on the real data; with EAO going from 0.289 on real data to 0.300 on generated data. This demonstrates our intuition that having great amounts of diverse data is very relevant when learning specialized deep features for TIR tracking. 
Finally, simultaneously using both real and generated data to fine-tune the network results in our best model. 
Moreover, training without distinguishing between the two types of data leads to better results, as opposed to a more complex two-stage fine-tuning process. 

% We use different thermal dataset to end-to-end fine-tuning the model. Firstly, we collect some real thermal videos and then make an IMDB to a $125\times 125$ patch size, where we denote as Real in Table~\ref{table: fine-tuning models}. And we also arrange the generated hallucination thermal videos deprived from two ways one is the pix2pix model~\cite{isola2017image} and the other is from cycle model~\cite{zhu2017unpaired}, where we denote data as GAN pix2pix and GAN cycle in Table~\ref{table: fine-tuning models}.

% Table~\ref{table: fine-tuning models} reports the performance results of VOT-TIR2016 challenge. they are ordered by the EAO metric, which unifies the robustness and accuracy of the trackers. Real+GAN pix2pix means that we mix the real TIR data and the pix2pix generated data together to jointly train the model. Real+GAN cycle means that we mix the real TIR data and the cycle generated data together to jointly train the model. Here we analysis that the mixed data contains more variance and diversity information, beneficial for training more robust model and strengthen the model to resist the noise.

\begin{figure}
    \centering
    \includegraphics[width=0.5\textwidth]{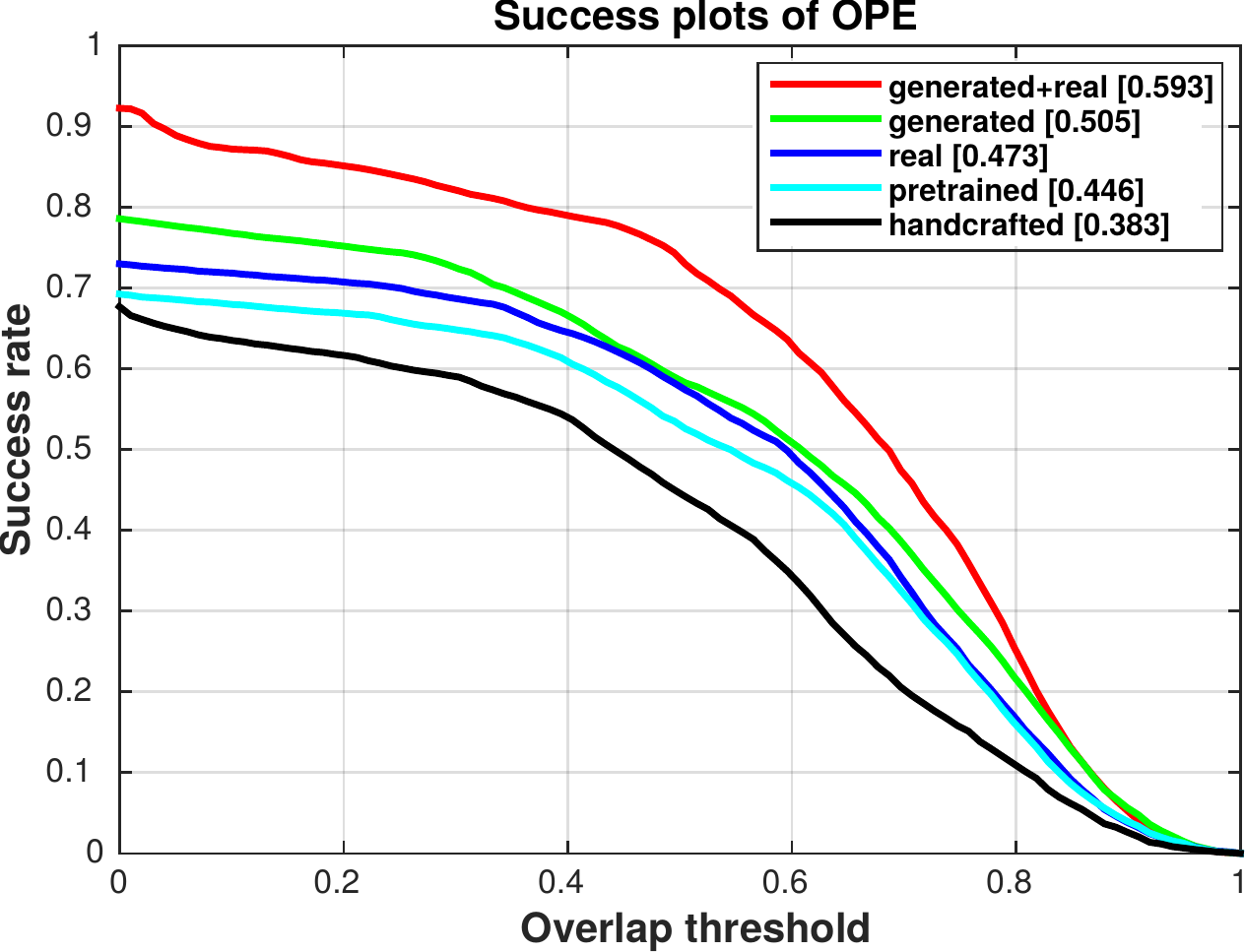}
	\caption{The success plot of one-pass evaluation (OPE) on the the VOT-TIR2017 benchmark~\cite{Kristan_2017_ICCV}. We show the AUC score of each tracker in the legend. The best results are obtained when using both real and generated data.}
    \label{fig:otb}
\end{figure}

We present results using the OPE evaluation metric in Fig.~\ref{fig:otb}. 
Also under this metric, handcrafted features show a clearly inferior performance compared to deep features. Simple pre-trained deep features obtain higher success rates, especially for mid-range overlap thresholds. Fine-tuning on real data gives the tracker a small boost, and when fine-tuning using our generated data, the performance is further improved. Finally, the best performance is achieved when fine-tuning using both types of data simultaneously.

\begin{figure}
    \centering
    \includegraphics[width=0.5\textwidth]{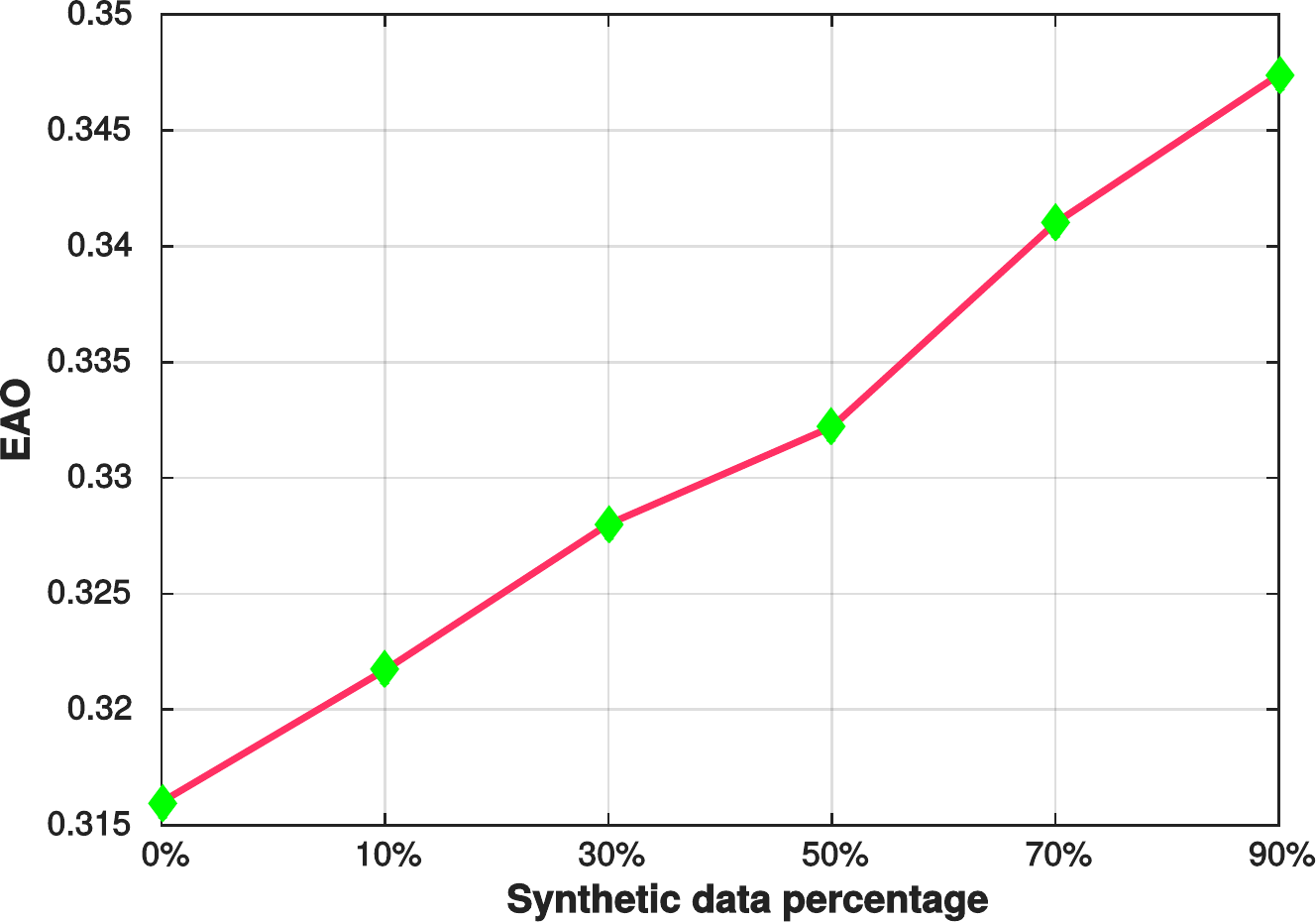}
	\caption{{Performance of our tracker (generated+real) on VOT-TIR2017~\cite{Kristan_2017_ICCV} for different percentages of synthetic data. The leftmost point indicates using only real data.}}
    \label{fig:percentage}
\end{figure}

{Finally, we analyze the performance of our generated + real tracker for different amounts of generated TIR data. 
Fig.~\ref{fig:percentage} shows EAO as a function of the percentage of synthetic TIR data in the total training set. 
Interestingly, increasing the amount of synthetic data monotonically improves the tracker performance.
The rightmost point, which corresponds to using all our generated data (90\% of the training set), does not seem to be saturated, and thus additional generated data could bring an even further performance boost.
}
% we can see that normally more various and diversity data improves the training quality and outperforms the model merely trained on real thermal data. Here we prefer the AUC score (success lot) is more accurate, as it considers the object's scale changes and is calculated by the area below the curve.

\subsection{Adding motion features}
As detailed in~\cite{7406435}, the use of handcrafted motion features can substantially improve tracking performance for TIR data. 
Following the implementation of the SRDCFir tracker~\cite{7406435}, we compute motion features by thresholding the absolute pixel-wise difference between the current and the previous frame.
We then use this motion mask as an extra feature channel.

Table~\ref{table:models} presents the results of our trained models when motion features are used alongside deep features.
We can see how motion features provide significant performance improvements to all models.
Furthermore, the conclusions drawn in the previous experiment still hold.
The models trained with generated data outperform both the pre-trained model and the model trained with real data only.
Finally, the model trained with a combination of generated and real data achieves an impressive performance, surpassing other methods.

A qualitative comparison baseline ECO (pretrained) and ours (generated) is shown in Fig.~\ref{fig:qualResECO}. In challenging cases (e.g. second row), the improved features learned through our generated TIR data lead to a tracking model that is accurate while robust to occlusion, scale change and out-of-plane rotation. 

% \setlength{\tabcolsep}{4pt}
% \begin{table}
%     \begin{center}        
%         \begin{tabular}{c|ccc}
%             \hline%\noalign{\smallskip}
% Tracker 		& EAO &A	&R\\ \hline 
% handcrafted		&0.361 &0.62 &1.12  \\
% pretrained 		&0.390 &0.65	&1.06 \\ 
% real  			&0.410 &0.63	&1.07\\
% generated 		&0.419&0.64	&0.99 \\ %\hline
% generated + real 	&\textbf{0.448}&\textbf{0.67}	&\textbf{0.89} \\   \hline        
%         \end{tabular}
%         \caption{Results of adding motion features to the fine-tuned models on VOT-TIR2017 dataset~\cite{Kristan_2017_ICCV}.}% \lichao{add the trained data?}\AG{compute missing row just in case}}
%         \label{table:motion}
%     \end{center}
% \end{table}

%Specifically in the IR data, it is very distinct between the foreground and background, so the motion changes between two images is very obvious to discriminate the object and background during tracking.
%Motion features are an one channel mask image which points the movement of the object and is obtained by the absoluting the difference between two adjacent frames under a threshold.
%The feature representation is shown as fig~\lichao{a figure show the motion feature}
%As SRDCFir utilized the motion features, it gets an jump improvement from SRDCF~\cite{danelljan2015learning} on TIR tracking. We integrate the motion mask into CNN features to improve our performance.

\subsection{State-of-the-art Comparison}
Here, we compare our best model with the three top TIR trackers in the VOT-TIR2017 challenge~\cite{Kristan_2017_ICCV}, i.e. DSLT~\cite{yu2017dense}, EBT~\cite{zhu2016beyond}, and SRDCFir~\cite{7406435}. 
We also include in our comparison the best CNN-based tracker in VOT-TIR2016, TCNN~\cite{nam2016modeling}. 
Additionally, we compare with recently introduced CF-based (CSRDCF)~\cite{lukezic2017discriminative} and spatial CF-based (CREST)~\cite{song2017crest} trackers. These trackers have shown excellent performance on VOT~\cite{VOT_TPAMI} and OTB~\cite{wu2015object} RGB datasets. 

Table~\ref{table:sota} shows the comparison of our best model (generated+real) including motion mask with the state-of-the-art methods in literature on the VOT-TIR2017 benchmark~\cite{Kristan_2017_ICCV}. Among the existing methods, SRDCFir and EBT achieve EAO scores of $0.364$ and $0.368$ respectively. An EAO score of $0.287$ is achieved by the TCNN tracker. The recently introduced CREST and CSRDCF trackers achieve EAO scores of $0.215$ and $0.248$ respectively. The current state-of-the-art on this dataset is the DSLT tracker with an EAO score of $0.401$. Our tracker significantly outperforms DSLT by setting a new state-of-the-art with an EAO score of $0.448$. Our approach also achieves superior performance in terms of accuracy and obtains second best results in terms of robustness. We further analyze the robustness of our tracker and found our approach to {have promising improvements with respect to robustness} in all videos except \textit{trees2}, compared to EBT. 

\setlength{\tabcolsep}{4pt}
\begin{table}
    \begin{center}        
        \begin{tabular}{c|ccc}
            \hline%\noalign{\smallskip}
            Tracker	& EAO 				& A 				& R \\ \hline            
            CREST 	& 0.215 			&0.56 		    	&4.13 \\   
            CSRDCF 	& 0.248 			&0.57				&3.49\\ \hline
            TCNN 	& 0.287 			&0.62 				&2.79 \\ 
            SRDCFir & 0.364 			&0.63	&1.10 \\
            EBT 	& 0.368 			&0.44 				&0.82 \\ 
            DSLT 	& {0.401}	&0.60 				&0.91 \\ \hline
            %Ours 	& \textbf{0.448}	&\textbf{0.67}  	&\underline{0.89}\\\hline 
            Ours 	& {\textbf{0.436}}	&{\textbf{0.65}}  	&{\textbf{0.80}}\\\hline 
        \end{tabular}
        \caption{Comparison with state-of-the-art trackers on VOT-TIR2017~\cite{Kristan_2017_ICCV}. Boldface indicates the best results. The results are reported in terms of expected average overlap (EAO), robustness (failure rate) and accuracy. Our proposed tracker significantly outperforms the state-of-the-art by achieving an EAO score of {$0.436$}. } % \AG{swap order of other trackers so DSLT is at the bottom, right above ours?}  and underline indicates second best
        \label{table:sota}
    \end{center}
\end{table}

%We significantly outperform all the considered trackers.Concretely, we obtain an EAO of 0.448, 0.047 points over the winner of the VOT-TIR2017 challenge, DSLT (0.401, according to~\cite{yu2017dense}), which makes our model trained using generated and real data the new state-of-the-art TIR tracker.
%We also substantially outperform CNN-based tracker TCNN as well as RGB trackers CSRDCF and CREST. We are only surpassed in terms of robustness by EBT~\cite{zhu2016beyond}, but our significantly higher accuracy (0.23 points) results in an overall superior tracker for the TIR modality. 

%\lichao{We analysis the robustness of our tracker compared with EBT~\cite{zhu2016beyond}, we found that in the majority of sequences, our tracker is robust except for video \textit{trees2}. We found EBT's robustness is higher just as its bounding box is big enough to bound the object. We also show our failure in Fig.~\ref{fig:qualResSota}}. \alertJW{Discuss}
\begin{figure*}
    \centering
    \includegraphics[width=\textwidth]{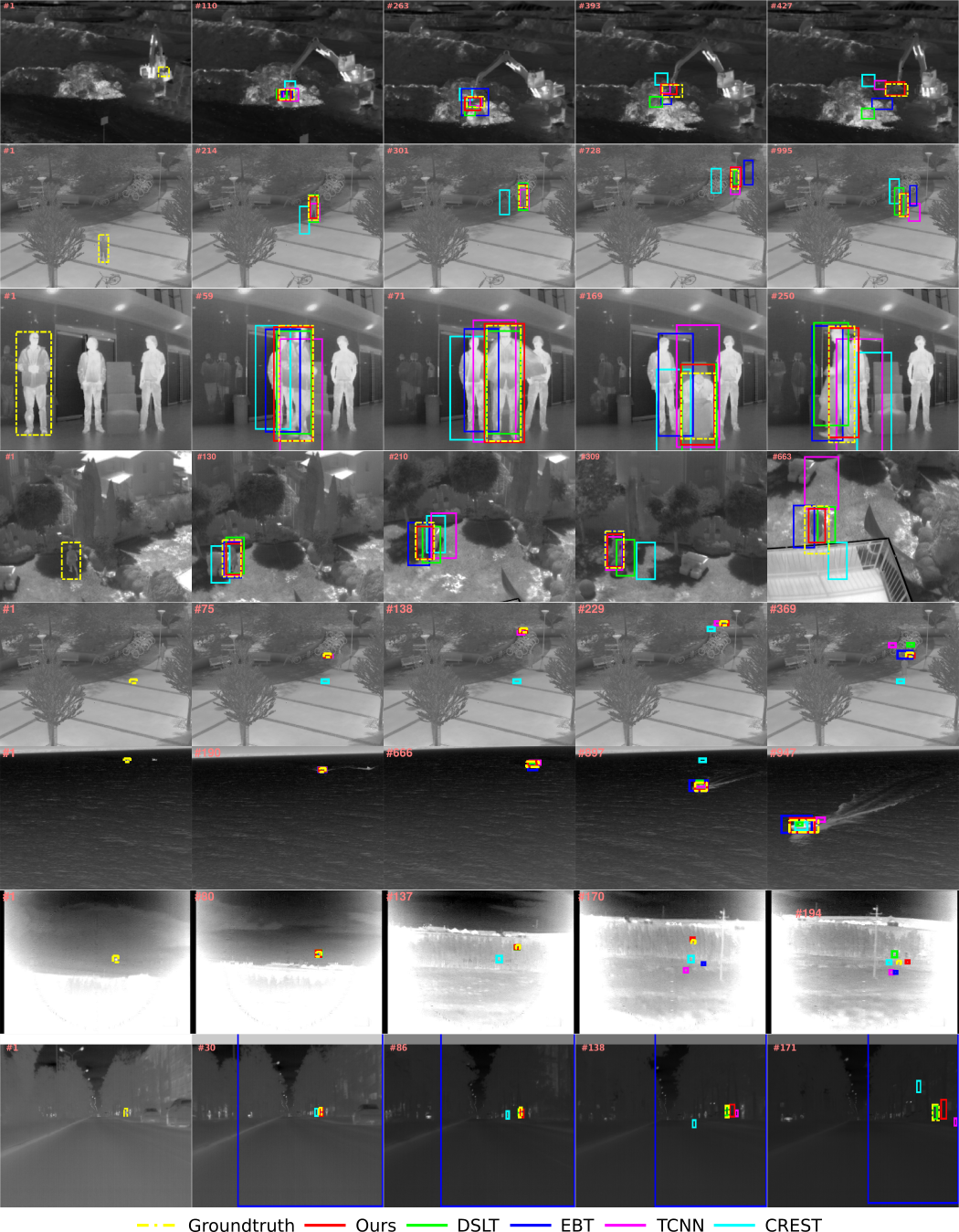}
    \caption{Qualitative comparison of our approach trained on generated and real data with state-of-the-art trackers, \textcolor{cyan}{CREST}, \textcolor{magenta}{TCNN}, \textcolor{blue}{EBT} and \textcolor{green}{DSLT} on the some challenging sequences, \textit{excavator}, \textit{jacket}, \textit{mixed\_distractors}, {\textit{garden}}, {\textit{quadrocopter2}}, {\textit{boat2}}, {\textit{bird}} and \textit{trees2} in VOT-TIR2017~\cite{Kristan_2017_ICCV}. Yellow dashed bounding box means \textcolor{yellow}{Groundtruth} and red solid bounding box is \textcolor{red}{Ours}. {The last two rows show failure cases of our tracker}.  }
    \label{fig:qualResSota}
\end{figure*}

Fig.~\ref{fig:qualResSota} shows a qualitative comparison of our tracker with state-of-the-art methods.
Our tracker follows the target object more accurately and is robust to challenging conditions such as scale change and occlusion.
Among existing methods, DSLT also provides improved tracking performance but struggles with accurate target localization. The proposed TIR-specialized deep features learned through abundant generated TIR data enable precise target localization, leading to superior tracking results. 
{The last two rows of the figure show two example cases in which our tracker fails.
In the first case, the object is rather tiny and lies on a cluttered background region, which increases the probability of confusing the tracked object with the background.
In the second case, there is a considerable scale change combined with heavy occlusion, leading to a poor estimation of the object extent and the corresponding tracking failure. 
}

\subsection{TIR Data Attributes Analysis}
\begin{figure*}
    \centering
    \includegraphics[width=\textwidth]{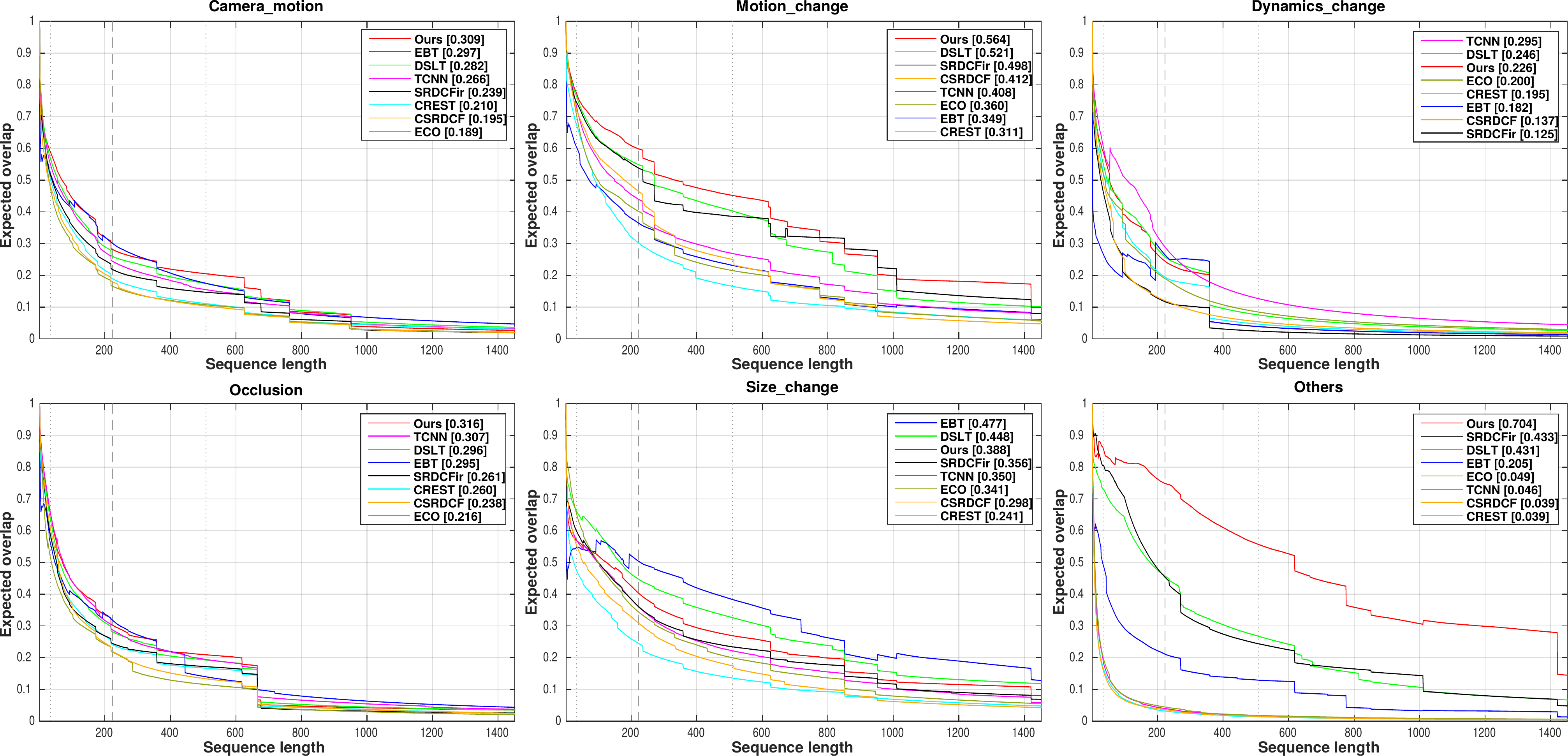}
    \caption{Attribute-based comparison of our trackers with state of-the-art methods on VOT-TIR2017 dataset. We show expected overlap measure 
for four attributes: camera motion, dynamics change, motion change, occlusion, size change, and others. Our trackers provide consistent improvements in case of camera motion, motion change, occlusion and others, compared to existing methods. }
    \label{fig:attrs}
\end{figure*}

In order to provide a more detailed analysis of the results, we present in Fig.~\ref{fig:attrs} the per-attribute performance comparison of our tracker and several state-of-the-art methods. The attributes are: camera motion, dynamics change, motion change, occlusion, size change, and others. Each attribute plot indicates the expected overlap for every tracker as a function of the sequence length, computed on a particular subset of videos annotated with the corresponding data attribute.
For most attributes, including the challenging scenarios of heavy camera motion, motion change, and occlusion, our proposed tracker outperforms state-of-the-art trackers. This consistent improvement on challenging attributes is likely due to specialized discriminative features, learned specifically for TIR tracking. In case of dynamics change, both TCNN and DSLT provide superior tracking performance. The TCNN tracker~\cite{nam2016modeling} can accurately match object proposals due to a tree structure encompassing multiple CNNs. The DSLT tracker~\cite{yu2017dense} also uses dense proposals and structural learning classifier. In case of size change, the EBT tracker~\cite{zhu2016beyond} and DSLT provide superior results.
In this attribute, our approach provides the third best results by outperforming trackers such as SRDCFir and TCNN. Overall, our approach achieves best results on 4 out of 6 attributes. 
%\lichao{We analysis in some cases, EBT predicts a huge big bounding box covering larger region than that of groundtruth like \#263 in video \textit{excavator}, and maybe even times of groundtruth such as \textit{trees2} in Fig.~\ref{fig:qualResSota}. While in this case even though the tracker itself does not provide an accurate prediction, it still obtains a higher score in the evaluation.}
%
% In motion\_change, we improve DSLT~\cite{yu2017dense} by $9.2\%$. As we end to end learn the model on TIR datasets and our models are adaptive abilities.
%In the case of videos with dynamics changes, however, TCNN and DSLT achieve higher performance than our tracker. TCNN~\cite{nam2016modeling} can accurately match object proposals due to a tree structure encompassing multiple CNNs, but it is also heavily time-consuming. DSLT~\cite{yu2017dense} also uses dense proposals and structural learning classifier.
%\AG{we need to find good reasons why those two outperform ours, and even suggest how we could solve this. From VOT-TIR: "  Not all cameras provide the full 16-bit range, instead, an adaptively changing 8-bit dynamics are sometimes used. Dynamics change indicates whether the dynamics is fixed during the sequence or not.JOost: martin and fahad say it is difficult}

\section{Conclusion}\label{sec:conclude}
In this paper, we have proposed a method to generate synthetic TIR data from RGB data. We use recent progress on image to image translation models for this purpose. The main advantage of this is that we can generate a large dataset of labeled TIR sequences. This dataset is far larger than datasets with real labeled sequences which are currently available for TIR tracking. These larger datasets allow us to perform end-to-end training for TIR features. To the best of our knowledge we are the first to train end-to-end features for TIR tracking. We show that our features trained on the synthetic data outperform other features for TIR tracking, including features which are computed by fine-tuning a network on real TIR sequences. In addition, we show that a combination of both real and generated data leads to a further improvement. Once we combine our feature with the motion feature we obtain state of the art results on the VOT-TIR2017. 

\section*{Acknowledgements}
This work was supported by TIN2016-79717-R, and the CHISTERA project M2CR (PCIN-2015-251) of the Spanish Ministry and the ACCIO agency and CERCA Programme / Generalitat de Catalunya. We also acknowledge the generous GPU support from NVIDIA.
%, and the EU Project CybSpeed MSCA-RISE-2017-777720

%\bibliographystyle{splncs}
\bibliographystyle{IEEEtran}
\bibliography{egbib}

\begin{thebibliography}{10}
\providecommand{\url}[1]{#1}
\csname url@samestyle\endcsname
\providecommand{\newblock}{\relax}
\providecommand{\bibinfo}[2]{#2}
\providecommand{\BIBentrySTDinterwordspacing}{\spaceskip=0pt\relax}
\providecommand{\BIBentryALTinterwordstretchfactor}{4}
\providecommand{\BIBentryALTinterwordspacing}{\spaceskip=\fontdimen2\font plus
\BIBentryALTinterwordstretchfactor\fontdimen3\font minus
  \fontdimen4\font\relax}
\providecommand{\BIBforeignlanguage}[2]{{%
\expandafter\ifx\csname l@#1\endcsname\relax
\typeout{** WARNING: IEEEtran.bst: No hyphenation pattern has been}%
\typeout{** loaded for the language `#1'. Using the pattern for}%
\typeout{** the default language instead.}%
\else
\language=\csname l@#1\endcsname
\fi
#2}}
\providecommand{\BIBdecl}{\relax}
\BIBdecl

\bibitem{emami2012role}
A.~Emami, F.~Dadgostar, A.~Bigdeli, and B.~C. Lovell, ``Role of spatiotemporal
  oriented energy features for robust visual tracking in video surveillance,''
  in \emph{Advanced Video and Signal-Based Surveillance (AVSS), 2012 IEEE Ninth
  International Conference on}.\hskip 1em plus 0.5em minus 0.4em\relax IEEE,
  2012, pp. 349--354.

\bibitem{renoust2016visual}
B.~Renoust, D.-D. Le, and S.~Satoh, ``Visual analytics of political networks
  from face-tracking of news video,'' \emph{IEEE Transactions on Multimedia},
  vol.~18, no.~11, pp. 2184--2195, 2016.

\bibitem{liu2012hand}
L.~Liu, J.~Xing, H.~Ai, and X.~Ruan, ``Hand posture recognition using finger
  geometric feature,'' in \emph{Pattern Recognition (ICPR), 2012 21st
  International Conference on}.\hskip 1em plus 0.5em minus 0.4em\relax IEEE,
  2012, pp. 565--568.

\bibitem{7406435}
K.~Alahari, A.~Berg, G.~Hager, J.~Ahlberg, M.~Kristan, J.~Matas, A.~Leonardis,
  L.~Cehovin, G.~Fernandez, T.~Vojir, G.~Nebehay, R.~Pflugfelder, A.~Lukezic,
  A.~Garcia-Martin, A.~Saffari, A.~Li, A.~S. Montero, B.~Zhao, C.~Schmid,
  D.~Chen, D.~Du, F.~S. Khan, F.~Porikli, G.~Zhu, G.~Zhu, H.~Lu, H.~Kieritz,
  H.~Li, H.~Qi, J.~chan Jeong, J.~il~Cho, J.-Y. Lee, J.~Zhu, J.~Li, J.~Feng,
  J.~Wang, J.-W. Kim, J.~Lang, J.~M. Martinez, K.~Xue, L.~Ma, L.~Ke, L.~Wen,
  L.~Bertinetto, M.~Danelljan, M.~Arens, M.~Tang, M.-C. Chang, O.~Miksik,
  P.~H.~S. Torr, R.~Martin-Nieto, R.~Laganière, S.~Hare, S.~Lyu, S.-C. Zhu,
  S.~Becker, S.~L. Hicks, S.~Golodetz, S.~Choi, T.~Wu, W.~Hübner, X.~Zhao,
  Y.~Hua, Y.~Li, Y.~Lu, Y.~Li, Z.~Yuan, and Z.~Hong, ``The thermal infrared
  visual object tracking vot-tir2015 challenge results,'' in \emph{2015 IEEE
  International Conference on Computer Vision Workshop (ICCVW)}, Dec 2015, pp.
  639--651.

\bibitem{felsberg2016thermal}
M.~Felsberg, M.~Kristan, J.~Matas, A.~Leonardis, R.~Pflugfelder, G.~H{\"a}ger,
  A.~Berg, A.~Eldesokey, J.~Ahlberg, L.~{\v{C}}ehovin \emph{et~al.}, ``The
  thermal infrared visual object tracking vot-tir2016 challenge results,'' in
  \emph{European Conference on Computer Vision}.\hskip 1em plus 0.5em minus
  0.4em\relax Springer, 2016, pp. 824--849.

\bibitem{Kristan_2017_ICCV}
M.~Kristan, A.~Leonardis, J.~Matas, M.~Felsberg, R.~Pflugfelder,
  L.~Cehovin~Zajc, T.~Vojir, G.~Hager, A.~Lukezic, A.~Eldesokey, and
  G.~Fernandez, ``The visual object tracking vot2017 challenge results,'' in
  \emph{The IEEE International Conference on Computer Vision (ICCV) Workshops},
  Oct 2017.

\bibitem{gade2014thermal}
R.~Gade and T.~B. Moeslund, ``Thermal cameras and applications: a survey,''
  \emph{Machine vision and applications}, vol.~25, no.~1, pp. 245--262, 2014.

\bibitem{bolme2010visual}
D.~S. Bolme, J.~R. Beveridge, B.~A. Draper, and Y.~M. Lui, ``Visual object
  tracking using adaptive correlation filters,'' in \emph{Computer Vision and
  Pattern Recognition (CVPR), 2010 IEEE Conference on}.\hskip 1em plus 0.5em
  minus 0.4em\relax IEEE, 2010, pp. 2544--2550.

\bibitem{henriques2015high}
J.~F. Henriques, R.~Caseiro, P.~Martins, and J.~Batista, ``High-speed tracking
  with kernelized correlation filters,'' \emph{IEEE Transactions on Pattern
  Analysis and Machine Intelligence}, vol.~37, no.~3, pp. 583--596, 2015.

\bibitem{danelljan2017eco}
M.~Danelljan, G.~Bhat, F.~S. Khan, and M.~Felsberg, ``Eco: efficient
  convolution operators for tracking,'' in \emph{Proceedings of the 2017 IEEE
  Conference on Computer Vision and Pattern Recognition (CVPR), Honolulu, HI,
  USA}, 2017, pp. 21--26.

\bibitem{wu2015object}
Y.~Wu, J.~Lim, and M.-H. Yang, ``Object tracking benchmark,'' \emph{IEEE
  Transactions on Pattern Analysis and Machine Intelligence}, vol.~37, no.~9,
  pp. 1834--1848, 2015.

\bibitem{VOT_TPAMI}
M.~Kristan, J.~Matas, A.~Leonardis, T.~Vojir, R.~Pflugfelder, G.~Fernandez,
  G.~Nebehay, F.~Porikli, and L.~\v{C}ehovin, ``A novel performance evaluation
  methodology for single-target trackers,'' \emph{IEEE Transactions on Pattern
  Analysis and Machine Intelligence}, vol.~38, no.~11, pp. 2137--2155, Nov
  2016.

\bibitem{mueller2016benchmark}
M.~Mueller, N.~Smith, and B.~Ghanem, ``A benchmark and simulator for uav
  tracking,'' in \emph{European Conference on Computer Vision}.\hskip 1em plus
  0.5em minus 0.4em\relax Springer, 2016, pp. 445--461.

\bibitem{danelljan2014adaptive}
M.~Danelljan, F.~S. Khan, M.~Felsberg, and J.~van~de Weijer, ``Adaptive color
  attributes for real-time visual tracking,'' in \emph{Computer Vision and
  Pattern Recognition (CVPR), 2014 IEEE Conference on}.\hskip 1em plus 0.5em
  minus 0.4em\relax IEEE, 2014, pp. 1090--1097.

\bibitem{danelljan2016adaptive}
M.~Danelljan, G.~Hager, F.~Shahbaz~Khan, and M.~Felsberg, ``Adaptive
  decontamination of the training set: A unified formulation for discriminative
  visual tracking,'' in \emph{Proceedings of the IEEE Conference on Computer
  Vision and Pattern Recognition}, 2016, pp. 1430--1438.

\bibitem{song2017crest}
Y.~Song, C.~Ma, L.~Gong, J.~Zhang, R.~W. Lau, and M.-H. Yang, ``Crest:
  Convolutional residual learning for visual tracking,'' in \emph{2017 IEEE
  International Conference on Computer Vision (ICCV)}.\hskip 1em plus 0.5em
  minus 0.4em\relax IEEE, 2017, pp. 2574--2583.

\bibitem{danelljan2015learning}
M.~Danelljan, G.~Hager, F.~Shahbaz~Khan, and M.~Felsberg, ``Learning spatially
  regularized correlation filters for visual tracking,'' in \emph{Proceedings
  of the IEEE International Conference on Computer Vision}, 2015, pp.
  4310--4318.

\bibitem{danelljan2014accurate}
M.~Danelljan, G.~H{\"a}ger, F.~Khan, and M.~Felsberg, ``Accurate scale
  estimation for robust visual tracking,'' in \emph{British Machine Vision
  Conference, Nottingham, September 1-5, 2014}.\hskip 1em plus 0.5em minus
  0.4em\relax BMVA Press, 2014.

\bibitem{krizhevsky2012imagenet}
A.~Krizhevsky, I.~Sutskever, and G.~E. Hinton, ``Imagenet classification with
  deep convolutional neural networks,'' in \emph{Advances in neural information
  processing systems}, 2012, pp. 1097--1105.

\bibitem{ma2015hierarchical}
C.~Ma, J.-B. Huang, X.~Yang, and M.-H. Yang, ``Hierarchical convolutional
  features for visual tracking,'' in \emph{Proceedings of the IEEE
  International Conference on Computer Vision}, 2015, pp. 3074--3082.

\bibitem{danelljan2016beyond}
M.~Danelljan, A.~Robinson, F.~S. Khan, and M.~Felsberg, ``Beyond correlation
  filters: Learning continuous convolution operators for visual tracking,'' in
  \emph{European Conference on Computer Vision}.\hskip 1em plus 0.5em minus
  0.4em\relax Springer, 2016, pp. 472--488.

\bibitem{russakovsky2015imagenet}
O.~Russakovsky, J.~Deng, H.~Su, J.~Krause, S.~Satheesh, S.~Ma, Z.~Huang,
  A.~Karpathy, A.~Khosla, M.~Bernstein \emph{et~al.}, ``Imagenet large scale
  visual recognition challenge,'' \emph{International Journal of Computer
  Vision}, vol. 115, no.~3, pp. 211--252, 2015.

\bibitem{valmadre2017end}
J.~Valmadre, L.~Bertinetto, J.~Henriques, A.~Vedaldi, and P.~H. Torr,
  ``End-to-end representation learning for correlation filter based tracking,''
  in \emph{Computer Vision and Pattern Recognition (CVPR), 2017 IEEE Conference
  on}.\hskip 1em plus 0.5em minus 0.4em\relax IEEE, 2017, pp. 5000--5008.

\bibitem{yu2017dense}
X.~Yu, Q.~Yu, Y.~Shang, and H.~Zhang, ``Dense structural learning for infrared
  object tracking at 200+ frames per second,'' \emph{Pattern Recognition
  Letters}, vol. 100, pp. 152--159, 2017.

\bibitem{dalal2005histograms}
N.~Dalal and B.~Triggs, ``Histograms of oriented gradients for human
  detection,'' in \emph{Computer Vision and Pattern Recognition, 2005. CVPR
  2005. IEEE Computer Society Conference on}, vol.~1.\hskip 1em plus 0.5em
  minus 0.4em\relax IEEE, 2005, pp. 886--893.

\bibitem{zhu2016beyond}
G.~Zhu, F.~Porikli, and H.~Li, ``Beyond local search: Tracking objects
  everywhere with instance-specific proposals,'' in \emph{Proceedings of the
  IEEE Conference on Computer Vision and Pattern Recognition}, 2016, pp.
  943--951.

\bibitem{goodfellow2014generative}
I.~Goodfellow, J.~Pouget-Abadie, M.~Mirza, B.~Xu, D.~Warde-Farley, S.~Ozair,
  A.~Courville, and Y.~Bengio, ``Generative adversarial nets,'' in
  \emph{Advances in neural information processing systems}, 2014, pp.
  2672--2680.

\bibitem{isola2017image}
P.~Isola, J.-Y. Zhu, T.~Zhou, and A.~A. Efros, ``Image-to-image translation
  with conditional adversarial networks,'' in \emph{Proceedings of the IEEE
  Conference on Computer Vision and Pattern Recognition}, 2017.

\bibitem{zhu2017unpaired}
J.-Y. Zhu, T.~Park, P.~Isola, and A.~A. Efros, ``Unpaired image-to-image
  translation using cycle-consistent adversarial networks,'' in
  \emph{Proceedings of the IEEE International Conference on Computer Vision},
  2017.

\bibitem{galoogahi2013multi}
H.~K. Galoogahi, T.~Sim, and S.~Lucey, ``Multi-channel correlation filters,''
  in \emph{Computer Vision (ICCV), 2013 IEEE International Conference
  on}.\hskip 1em plus 0.5em minus 0.4em\relax IEEE, 2013, pp. 3072--3079.

\bibitem{van2009learning}
J.~Van De~Weijer, C.~Schmid, J.~Verbeek, and D.~Larlus, ``Learning color names
  for real-world applications,'' \emph{Image Processing, IEEE Transactions on},
  vol.~18, no.~7, pp. 1512--1523, 2009.

\bibitem{cf_ca_tracking}
M.~Mueller, N.~Smith, and B.~Ghanem, ``Context-aware correlation filter
  tracking,'' in \emph{Proc. of the IEEE Conference on Computer Vision and
  Pattern Recognition (CVPR)}, 2017.

\bibitem{ma2015long}
C.~Ma, X.~Yang, C.~Zhang, and M.-H. Yang, ``Long-term correlation tracking,''
  in \emph{Computer Vision and Pattern Recognition (CVPR), 2015 IEEE Conference
  on}.\hskip 1em plus 0.5em minus 0.4em\relax IEEE, 2015, pp. 5388--5396.

\bibitem{lukezic2017discriminative}
A.~Lukezic, T.~Voj{\'\i}r, L.~C. Zajc, J.~Matas, and M.~Kristan,
  ``Discriminative correlation filter with channel and spatial reliability,''
  in \emph{Proceedings of the IEEE Conference on Computer Vision and Pattern
  Recognition}, vol.~2, 2017.

\bibitem{kiani2017learning}
H.~Kiani~Galoogahi, A.~Fagg, and S.~Lucey, ``Learning background-aware
  correlation filters for visual tracking,'' in \emph{Proceedings of the IEEE
  Conference on Computer Vision and Pattern Recognition}, 2017, pp. 1135--1143.

\bibitem{danelljan2015convolutional}
M.~Danelljan, G.~Hager, F.~Shahbaz~Khan, and M.~Felsberg, ``Convolutional
  features for correlation filter based visual tracking,'' in \emph{Proceedings
  of the IEEE International Conference on Computer Vision Workshops}, 2015, pp.
  58--66.

\bibitem{yu2017online}
X.~Yu and Q.~Yu, ``Online structural learning with dense samples and a
  weighting kernel,'' \emph{Pattern Recognition Letters}, 2017.

\bibitem{li2016learning}
C.~Li, H.~Cheng, S.~Hu, X.~Liu, J.~Tang, and L.~Lin, ``Learning collaborative
  sparse representation for grayscale-thermal tracking,'' \emph{IEEE
  Transactions on Image Processing}, vol.~25, no.~12, pp. 5743--5756, 2016.

\bibitem{li2017grayscale}
C.~Li, X.~Sun, X.~Wang, L.~Zhang, and J.~Tang, ``Grayscale-thermal object
  tracking via multitask laplacian sparse representation,'' \emph{IEEE
  Transactions on Systems, Man, and Cybernetics: Systems}, vol.~47, no.~4, pp.
  673--681, 2017.

\bibitem{hoffman2016learning}
J.~Hoffman, S.~Gupta, and T.~Darrell, ``Learning with side information through
  modality hallucination,'' in \emph{Proceedings of the IEEE Conference on
  Computer Vision and Pattern Recognition}, 2016, pp. 826--834.

\bibitem{xu2017learning}
D.~Xu, W.~Ouyang, E.~Ricci, X.~Wang, and N.~Sebe, ``Learning cross-modal deep
  representations for robust pedestrian detection,'' in \emph{Computer Vision
  and Pattern Recognition (CVPR)}.\hskip 1em plus 0.5em minus 0.4em\relax IEEE,
  2017.

\bibitem{denton2015deep}
E.~L. Denton, S.~Chintala, R.~Fergus \emph{et~al.}, ``Deep generative image
  models using a laplacian pyramid of adversarial networks,'' in \emph{Advances
  in neural information processing systems}, 2015, pp. 1486--1494.

\bibitem{perarnau2016invertible}
G.~Perarnau, J.~van~de Weijer, B.~Raducanu, and J.~M. {\'A}lvarez, ``Invertible
  conditional gans for image editing,'' in \emph{NIPS 2016 Workshop on
  Adversarial Training}, 2016.

\bibitem{salimans2016improved}
T.~Salimans, I.~Goodfellow, W.~Zaremba, V.~Cheung, A.~Radford, and X.~Chen,
  ``Improved techniques for training gans,'' in \emph{Advances in Neural
  Information Processing Systems}, 2016, pp. 2234--2242.

\bibitem{mirza2014conditional}
M.~Mirza and S.~Osindero, ``Conditional generative adversarial nets,''
  \emph{arXiv preprint arXiv:1411.1784}, 2014.

\bibitem{zhang2016colorful}
R.~Zhang, P.~Isola, and A.~A. Efros, ``Colorful image colorization,'' in
  \emph{European Conference on Computer Vision}.\hskip 1em plus 0.5em minus
  0.4em\relax Springer, 2016, pp. 649--666.

\bibitem{ronneberger2015u}
O.~Ronneberger, P.~Fischer, and T.~Brox, ``U-net: Convolutional networks for
  biomedical image segmentation,'' in \emph{International Conference on Medical
  image computing and computer-assisted intervention}.\hskip 1em plus 0.5em
  minus 0.4em\relax Springer, 2015, pp. 234--241.

\bibitem{wu2014thermal}
Z.~Wu, N.~Fuller, D.~Theriault, and M.~Betke, ``A thermal infrared video
  benchmark for visual analysis,'' in \emph{Computer Vision and Pattern
  Recognition Workshops (CVPRW), 2014 IEEE Conference on}.\hskip 1em plus 0.5em
  minus 0.4em\relax IEEE, 2014, pp. 201--208.

\bibitem{song2017depth}
X.~Song, L.~Herranz, and S.~Jiang, ``Depth cnns for rgb-d scene recognition:
  Learning from scratch better than transferring from rgb-cnns.'' 2017.

\bibitem{hwang2013multispectral}
S.~Hwang, J.~Park, N.~Kim, Y.~Choi, and I.~S. Kweon, ``Multispectral pedestrian
  detection: Benchmark dataset and baseline,'' \emph{Integrated Comput.-Aided
  Eng}, vol.~20, pp. 347--360, 2013.

\bibitem{wang2017dcfnet}
Q.~Wang, J.~Gao, J.~Xing, M.~Zhang, and W.~Hu, ``Dcfnet: Discriminant
  correlation filters network for visual tracking,'' \emph{arXiv preprint
  arXiv:1704.04057}, 2017.

\bibitem{gundogdu2017good}
E.~Gundogdu and A.~A. Alatan, ``Good features to correlate for visual
  tracking,'' \emph{arXiv preprint arXiv:1704.06326}, 2017.

\bibitem{nocedal2006numerical}
J.~Nocedal and S.~J. Wright, ``Numerical optimization 2nd,'' 2006.

\bibitem{li2016precomputed}
C.~Li and M.~Wand, ``Precomputed real-time texture synthesis with markovian
  generative adversarial networks,'' in \emph{European Conference on Computer
  Vision}.\hskip 1em plus 0.5em minus 0.4em\relax Springer, 2016, pp. 702--716.

\bibitem{gonzalez2016pedestrian}
A.~Gonz{\'a}lez, Z.~Fang, Y.~Socarras, J.~Serrat, D.~V{\'a}zquez, J.~Xu, and
  A.~M. L{\'o}pez, ``Pedestrian detection at day/night time with visible and
  fir cameras: A comparison,'' \emph{Sensors}, vol.~16, no.~6, p. 820, 2016.

\bibitem{davis2005two}
J.~W. Davis and M.~A. Keck, ``A two-stage template approach to person detection
  in thermal imagery,'' in \emph{Application of Computer Vision, 2005.
  WACV/MOTIONS'05 Volume 1. Seventh IEEE Workshops on}, vol.~1.\hskip 1em plus
  0.5em minus 0.4em\relax IEEE, 2005, pp. 364--369.

\bibitem{palmero2016multi}
C.~Palmero, A.~Clap{\'e}s, C.~Bahnsen, A.~M{\o}gelmose, T.~B. Moeslund, and
  S.~Escalera, ``Multi-modal rgb--depth--thermal human body segmentation,''
  \emph{International Journal of Computer Vision}, vol. 118, no.~2, pp.
  217--239, 2016.

\bibitem{bilodeau2014thermal}
G.-A. Bilodeau, A.~Torabi, P.-L. St-Charles, and D.~Riahi, ``Thermal--visible
  registration of human silhouettes: A similarity measure performance
  evaluation,'' \emph{Infrared Physics \& Technology}, vol.~64, pp. 79--86,
  2014.

\bibitem{torabi2012iterative}
A.~Torabi, G.~Mass{\'e}, and G.-A. Bilodeau, ``An iterative integrated
  framework for thermal--visible image registration, sensor fusion, and people
  tracking for video surveillance applications,'' \emph{Computer Vision and
  Image Understanding}, vol. 116, no.~2, pp. 210--221, 2012.

\bibitem{kristan2016VOT}
M.~Kristan, R.~Pflugfelder, A.~Leonardis, J.~Matas, F.~Porikli, L.~Cehovin,
  G.~Nebehay, G.~Fernandez, T.~Vojir, A.~Gatt \emph{et~al.}, ``The visual
  object tracking vot2016 challenge results,'' in \emph{Computer Vision
  Workshops (ECCVW), European Conference on}, 2016.

\bibitem{portmann2014people}
J.~Portmann, S.~Lynen, M.~Chli, and R.~Siegwart, ``People detection and
  tracking from aerial thermal views,'' in \emph{Robotics and Automation
  (ICRA), 2014 IEEE International Conference on}.\hskip 1em plus 0.5em minus
  0.4em\relax IEEE, 2014, pp. 1794--1800.

\bibitem{gade2013long}
R.~Gade, A.~J{\o}rgensen, and T.~B. Moeslund, ``Long-term occupancy analysis
  using graph-based optimisation in thermal imagery,'' in \emph{Computer Vision
  and Pattern Recognition (CVPR), 2013 IEEE Conference on}.\hskip 1em plus
  0.5em minus 0.4em\relax IEEE, 2013, pp. 3698--3705.

\bibitem{gao2016infar}
C.~Gao, Y.~Du, J.~Liu, J.~Lv, L.~Yang, D.~Meng, and A.~G. Hauptmann, ``Infar
  dataset: Infrared action recognition at different times,''
  \emph{Neurocomputing}, vol. 212, pp. 36--47, 2016.

\bibitem{vcehovin2016visual}
L.~{\v{C}}ehovin, A.~Leonardis, and M.~Kristan, ``Visual object tracking
  performance measures revisited,'' \emph{IEEE Transactions on Image
  Processing}, vol.~25, no.~3, pp. 1261--1274, 2016.

\bibitem{chatfield2014return}
K.~Chatfield, K.~Simonyan, A.~Vedaldi, and A.~Zisserman, ``Return of the devil
  in the details: Delving deep into convolutional nets,'' \emph{arXiv preprint
  arXiv:1405.3531}, 2014.

\bibitem{he2016deep}
K.~He, X.~Zhang, S.~Ren, and J.~Sun, ``Deep residual learning for image
  recognition,'' in \emph{Proceedings of the IEEE conference on computer vision
  and pattern recognition}, 2016, pp. 770--778.

\bibitem{ecocode}
\url{https://github.com/martin-danelljan/ECO}.

\bibitem{nam2016learning}
H.~Nam and B.~Han, ``Learning multi-domain convolutional neural networks for
  visual tracking,'' in \emph{Proceedings of the IEEE Conference on Computer
  Vision and Pattern Recognition}, 2016, pp. 4293--4302.

\bibitem{park2018meta}
E.~Park and A.~C. Berg, ``Meta-tracker: Fast and robust online adaptation for
  visual object trackers,'' \emph{arXiv preprint arXiv:1801.03049}, 2018.

\bibitem{nam2016modeling}
H.~Nam, M.~Baek, and B.~Han, ``Modeling and propagating cnns in a tree
  structure for visual tracking,'' \emph{arXiv preprint arXiv:1608.07242},
  2016.

\end{thebibliography}
\end{document}